\documentclass{article} 
\usepackage{iclr2021_conference,times}

\usepackage{amsmath,amsfonts,bm}

\def\eqref#1{equation~\ref{#1}}

\def\1{\bm{1}}

\DeclareMathAlphabet{\mathsfit}{\encodingdefault}{\sfdefault}{m}{sl}
\SetMathAlphabet{\mathsfit}{bold}{\encodingdefault}{\sfdefault}{bx}{n}

\DeclareMathOperator{\sign}{sign}

\usepackage{amsmath}
\usepackage{amssymb}
\usepackage{times}
\usepackage{bm}
\usepackage{xargs}
\usepackage{multirow}
\usepackage{amsfonts}
\usepackage{color}
\usepackage{booktabs}

\newcommandx\includeImageLineWidth[2][1=1.0]{\includegraphics[width=#1\linewidth]{#2}}

\newcommand{\bq}{{\bm{q}}}

\newcommand{\bs}{{\bm{s}}}
\newcommand{\bt}{{\bm{t}}}

\newcommand{\bw}{{\bm{w}}}
\newcommand{\bx}{{\bm{x}}}
\newcommand{\by}{{\bm{y}}}

\newcommand{\bX}{{\bm{X}}}
\newcommand{\bY}{{\bm{Y}}}

\usepackage{array}
\newcommand{\PreserveBackslash}[1]{\let\temp=\\#1\let\\=\temp}
\newcolumntype{C}[1]{>{\PreserveBackslash\centering}p{#1}}
\newcolumntype{R}[1]{>{\PreserveBackslash\raggedleft}p{#1}}
\newcolumntype{L}[1]{>{\PreserveBackslash\raggedright}p{#1}}

\usepackage[colorlinks,citecolor=teal]{hyperref}
\usepackage{url}
\usepackage{booktabs}       
\usepackage{amsfonts}       
\usepackage{nicefrac}       
\usepackage{microtype}      
\usepackage{algorithm}
\usepackage{algorithmic}
\usepackage{graphicx}
\usepackage{subfigure}
\usepackage{subfiles}
\usepackage{pifont}
\usepackage{enumitem}
\usepackage{wrapfig}

\title{Growing Efficient Deep Networks\\by Structured Continuous Sparsification}

\author{Xin Yuan \\
University of Chicago\\
\texttt{yuanx@uchicago.edu} \\
\And
Pedro Savarese \\
TTI-Chicago \\
\texttt{savarese@ttic.edu} \\
\And
Michael Maire \\
University of Chicago \\
\texttt{mmaire@uchicago.edu}
}

\newcommand{\cmark}{\ding{51}}%
\newcommand{\xmark}{\ding{55}}%

\iclrfinalcopy 
\begin{document}

\maketitle

\begin{abstract}
We develop an approach to growing deep network architectures over the course of training, driven by a principled combination of accuracy and sparsity objectives.  Unlike existing pruning or architecture search techniques that operate on full-sized models or supernet architectures, our method can start from a small, simple seed architecture and dynamically grow and prune both layers and filters.  By combining a continuous relaxation of discrete network structure optimization with a scheme for sampling sparse subnetworks, we produce compact, pruned networks, while also drastically reducing the computational expense of training.  For example, we achieve $49.7\%$ inference FLOPs and $47.4\%$ training FLOPs savings compared to a baseline ResNet-50 on ImageNet, while maintaining $75.2\%$ top-1 accuracy --- all without any dedicated fine-tuning stage.  Experiments across CIFAR, ImageNet, PASCAL VOC, and Penn Treebank, with convolutional networks for image classification and semantic segmentation, and recurrent networks for language modeling, demonstrate that we both train faster and produce more efficient networks than competing architecture pruning or search methods.

\end{abstract}

\section{Introduction}
\label{sec:intro}

Deep neural networks are the dominant approach to a variety of machine learning
tasks, including
image classification~\citep{
krizhevsky2012imagenet,
DBLP:journals/corr/SimonyanZ14a},
object detection~\citep{
DBLP:conf/iccv/Girshick15,
DBLP:conf/eccv/LiuAESRFB16},
semantic segmentation~\citep{
DBLP:conf/cvpr/LongSD15,
DBLP:journals/corr/ChenPSA17}
and language modeling~\citep{
DBLP:journals/corr/ZarembaSV14,
vaswani2017attention,
DBLP:conf/naacl/DevlinCLT19}.
Modern neural networks are overparameterized and training larger networks
usually yields improved generalization accuracy.  Recent
work~\citep{
DBLP:conf/cvpr/HeZRS16,
DBLP:conf/bmvc/ZagoruykoK16,
DBLP:conf/cvpr/HuangLMW17}
illustrates this trend through increasing \emph{depth} and \emph{width} of
convolutional neural networks (CNNs).  Yet, training is compute-intensive, and
real-world deployments are often limited by parameter and compute budgets.

Neural architecture search (NAS)~\citep{
DBLP:conf/iclr/ZophL17,
liu2018darts,
luo2018neural,
DBLP:conf/icml/PhamGZLD18,
implicitrecurrent} and
model pruning~\citep{
DBLP:journals/corr/HanMD15,
DBLP:journals/corr/HanPTD15,
DBLP:conf/nips/GuoYC16}
methods aim to reduce these burdens.  NAS addresses an issue that further
compounds training cost: the enormous space of possible network architectures.
While hand-tuning architectural details, such as the connection structure
of convolutional layers, can improve performance~\citep{
DBLP:journals/corr/IandolaMAHDK16,
sifre2014rigid,DBLP:conf/cvpr/Chollet17,
DBLP:journals/corr/HowardZCKWWAA17,
DBLP:journals/corr/ZhangZLS17,
DBLP:journals/corr/abs-1711-09224},
a principled way of deriving such designs remains elusive.  NAS methods aim to
automate exploration of possible architectures, producing an efficient design
for a target task under practical resource constraints.  However, during
training, most NAS methods operate on a large \emph{supernet} architecture,
which encompasses candidate components beyond those that are eventually
selected for inclusion in the resulting network~\citep{
DBLP:conf/iclr/ZophL17,
liu2018darts,
luo2018neural,
DBLP:conf/icml/PhamGZLD18,
implicitrecurrent}.
Consequently, NAS-based training may typically be more thorough, but
more computationally expensive, than training a single hand-designed
architecture.

Model pruning techniques similarly focus on improving the resource efficiency
of neural networks during inference, at the possible expense of increased
training cost.  Common strategies aim to generate a lighter version of a given
network architecture by removing individual weights~\citep{
DBLP:journals/corr/HanPTD15,
DBLP:journals/corr/HanMD15,
DBLP:conf/icml/MolchanovAV17}
or structured parameter sets~\citep{
DBLP:conf/iclr/0022KDSG17,
DBLP:conf/ijcai/HeKDFY18,DBLP:conf/iccv/LuoWL17}.
However, the majority of these methods train a full-sized model prior to
pruning and, after pruning, utilize additional fine-tuning phases in order to
maintain accuracy.  \cite{DBLP:conf/nips/HubaraCSEB16} and
\cite{DBLP:conf/eccv/RastegariORF16} propose the use of binary weights and
activations, allowing inference to benefit from reduced storage costs and
efficient computation through bit-counting operations.  Yet, training still
involves tracking high-precision weights alongside lower-precision
approximations.

\begin{figure}[t]
   \vspace{-2.0em}
   \begin{minipage}[b]{\linewidth}
   \subfigure[Growing CNN Layers and Filters]{
      \includegraphics[height=0.20\columnwidth]{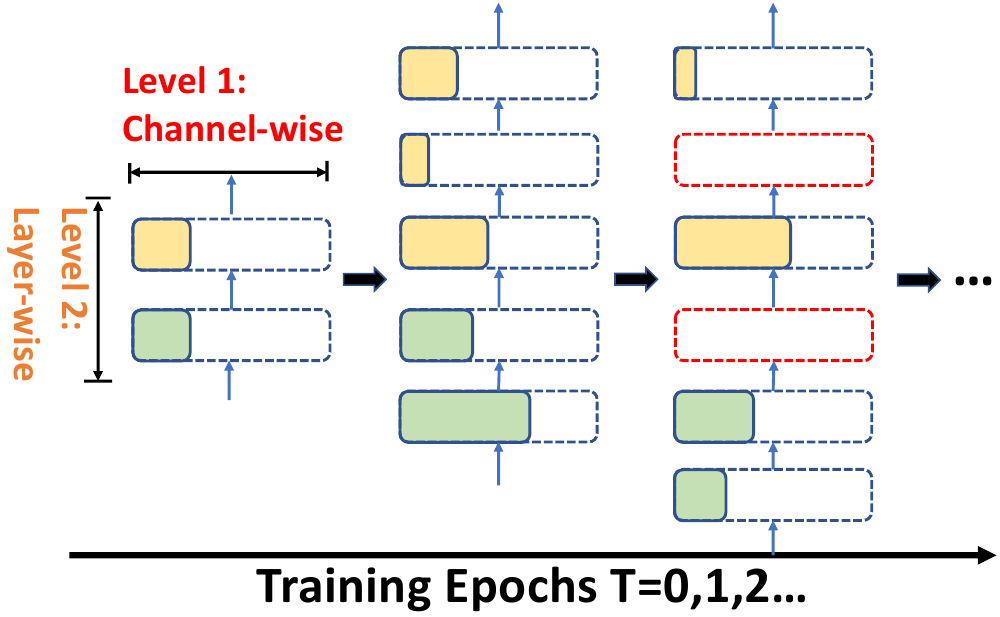}
   }
   \hfill
   \subfigure[Epoch-wise Training Cost]{
      \includegraphics[height=0.20\columnwidth]{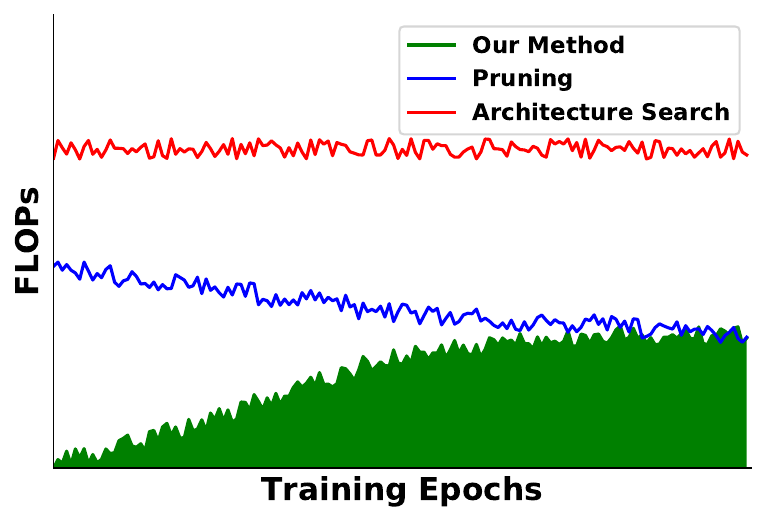}
   }
   \hfill
   \subfigure[Total Training Cost]{
      \includegraphics[height=0.20\columnwidth]{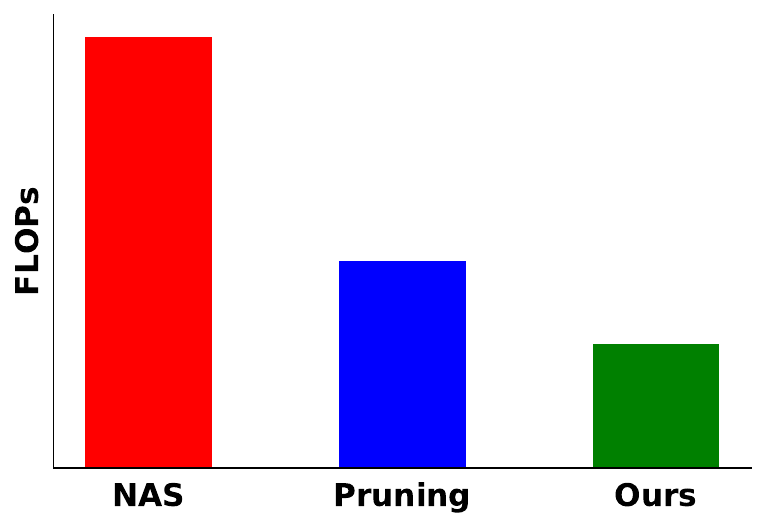}
   }
   \vspace{-0.5em}
   \caption{\footnotesize{%
      \textbf{Growing Networks during Training.}
      We define an architecture configuration space and simultaneously adapt
      network structure and weights.
      \textbf{(a)}
      Applying our approach to CNNs, we maintain auxiliary variables that
      determine how to grow and prune both filters (\emph{i.e.}~channel-wise)
      and layers, subject to practical resource constraints.
      \textbf{(b)}
      By starting with a small network and growing its size, we utilize fewer
      resources in early training epochs, compared to pruning or NAS methods.
      \textbf{(c)}
      Consequently, our method significantly reduces the total computational
      cost of training, while delivering trained networks of comparable or
      better size and accuracy.
   }}
   \label{fig:motivation}
   \end{minipage}
   \vspace{-1.0em}
\end{figure}

\begin{figure}[ht]
   \vspace{-2.0em}
   \hfill
   \subfigure[Architectural Configuration Space for CNNs]{
      \label{fig:growspace}
      \includegraphics[height=0.348\columnwidth]{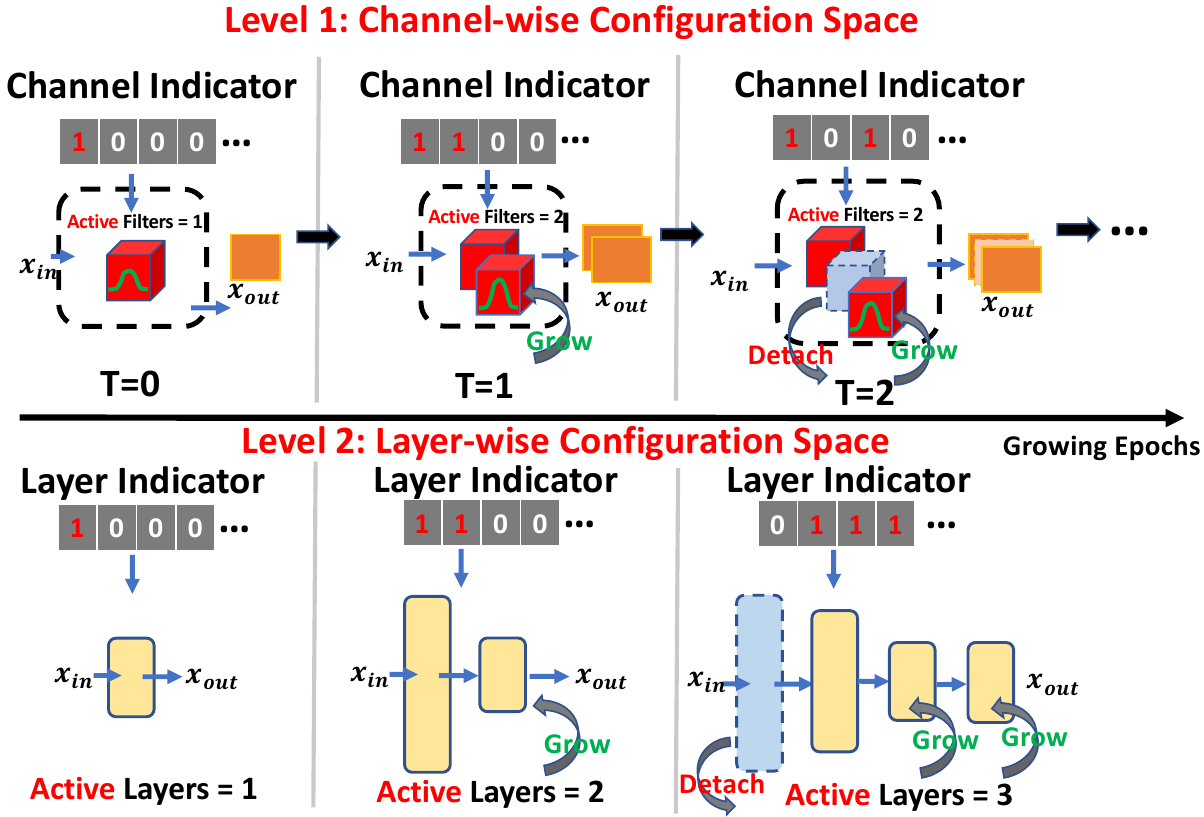}
   }
   \hfill
   \subfigure[Optimization with Structured Continuation]{
      \label{fig:growprocedure}
      \includegraphics[height=0.348\columnwidth]{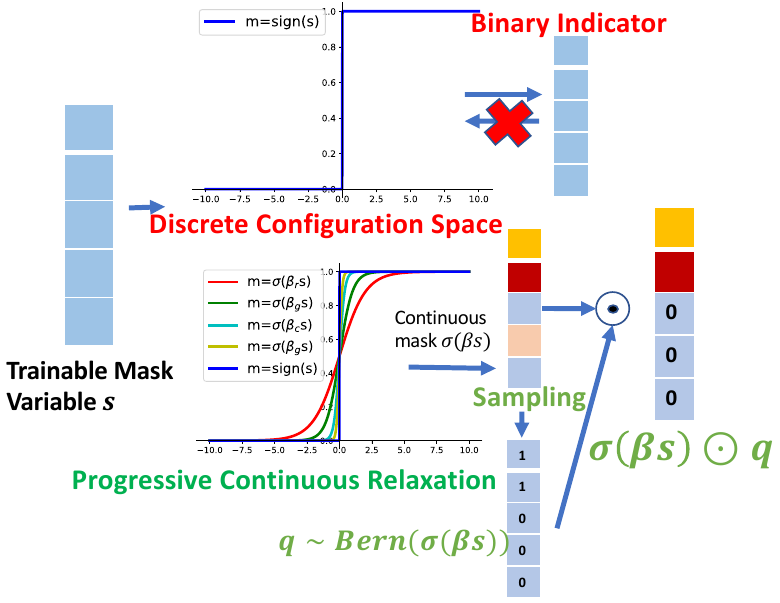}
   }
   \hfill
   \vspace{-1.0em}
   \caption{\footnotesize{%
      \textbf{Technical Framework.}
      \textbf{(a)}
      We periodically restructure a CNN by querying binary indicators that
      define a two-level configuration space for filters and layers.
      \textbf{(b)}
      To make optimization feasible while growing networks, we derive these
      binary indicators from trainable continuous mask variables.  We employ a
      structured extension of continuous sparsification~\citep{%
      DBLP:conf/nips/SavareseSM20}, combined with sampling.  Binary
      stochastic auxiliary variables $q$, sampled according to $\sigma(\beta s)$,
      generate the discrete components active at a particular time.
   }}
   \label{fig:method_frame}
   \vspace{-1.0em}
\end{figure}

We take a unified view of pruning and architecture search, regarding both as
acting on a configuration space, and propose a method to dynamically grow deep
networks by continuously reconfiguring their architecture during training.
Our approach not only produces models with efficient inference characteristics,
but also reduces the computational cost of training; see
Figure~\ref{fig:motivation}.  Rather than starting with a full-sized network
or a supernet, we start from simple seed networks and progressively adjust
(grown and prune) them.  Specifically, we parameterize an architectural
configuration space with indicator variables governing addition or removal of
structural components.  Figure~\ref{fig:growspace} shows an example, in the
form of a two-level configuration space for CNN layers and filters.  We enable
learning of indicator values (and thereby, architectural structure) via
combining a continuous relaxation with binary sampling, as illustrated in
Figure~\ref{fig:growprocedure}.  A per-component temperature parameter ensures
that long-lived structures are eventually baked into the network's discrete
architectural configuration.

While the recently proposed AutoGrow~\citep{DBLP:conf/kdd/Wen0CL20} also seeks
to grow networks over the course of training, our technical approach differs
substantially and leads to significant practical advantages.  At a technical
level, AutoGrow implements an architecture search procedure over a predefined
modular structure, subject to hand-crafted, accuracy-driven growing and
stopping policies.  In contrast, we parameterize architectural configurations
and utilize stochastic gradient descent to learn the auxiliary variables that
specify structural components, while simultaneously training the weights within
those components.  Our unique technical approach yields the following
advantages:
\begin{itemize}[leftmargin=.15in]
   \item{%
      \textbf{Fast Training by Growing:}
      Training is a unified procedure, from which one can request a network
      structure and associated weights at any time.  Unlike AutoGrow and the
      majority of pruning techniques, fine-tuning to optimize weights in a
      discovered architecture is optional.  We achieve excellent results
      even without any fine-tuning stage.
   }%
   \item{%
      \textbf{Principled Approach via Learning by Continuation + Sampling:}
      We formulate our approach in the spirit of learning by continuation
      methods, which relax a discrete optimization problem to an increasingly
      stiff continuous approximation.  Critically, we introduce an additional
      sampling step to this strategy.  From this combination, we gain the
      flexibility of exploring a supernet architecture, but the computational
      efficiency of only actually training a much smaller active subnetwork.
   }%
   \item{%
      \textbf{Budget-Aware Optimization Objectives:}
      The parameters governing our architectural configuration are themselves
      updated via gradient decent.  We have flexibility to formulate a variety
      of resource-sensitive losses, such as counting total FLOPs, in terms of
      these parameters.
   }%
   \item{%
      \textbf{Broad Applicability:}
      Though we use progressive growth of CNNs in width and depth as a
      motivating example, our technique applies to virtually any neural
      architecture.  One has flexibility in how to parameterize the
      architecture configuration space.  We also show results with LSTMs.
   }%
\end{itemize}
We demonstrate these advantages while comparing to recent NAS and pruning
methods through extensive experiments on classification, semantic segmentation,
and word-level language modeling.

\section{Related Work}
\label{sec:related}

\vspace{-0.5em}
\textbf{Network Pruning.}
Pruning methods can be split into two groups: those pruning individual weights
and those pruning structured components.  Individual weight-based pruning
methods vary on the removal criteria.  For example,
\cite{DBLP:journals/corr/HanPTD15} propose to prune network weights with small
magnitude, and subsequently quantize those remaining~\citep{
DBLP:journals/corr/HanMD15}.
\cite{DBLP:journals/corr/abs-1712-01312} learn sparse networks by
approximating $\ell_0$-regularization with a stochastic reparameterization.
However, sparse weights alone often only lead to speedups on
dedicated hardware with supporting libraries.

In structured methods, pruning is applied at the level of neurons, channels,
or even layers.  For example, L1-pruning~\citep{DBLP:conf/iclr/0022KDSG17}
removes channels based on the norm of their filters.
\cite{DBLP:conf/ijcai/HeKDFY18} use group sparsity to smooth the pruning
process after training.
%
%
MorphNet~\citep{DBLP:conf/cvpr/GordonENCWYC18} regularizes weights towards zero
until they are small enough such that the corresponding output channels are
marked for removal from the network.
Intrinsic Structured Sparsity (ISS)~\citep{DBLP:conf/iclr/WenHRZWLHC018} works
on LSTMs~\citep{LSTM} by collectively removing the columns and rows of the
weight matrices via group LASSO.
Although structured pruning methods and our algorithm share the same spirit of
generating efficient models, we gain training cost savings by growing
networks from small initial architectures instead of pruning full-sized ones.

\textbf{Neural Architecture Search.}
NAS methods have greatly improved the performance achieved by small network
models.  Pioneering NAS approaches use reinforcement learning~\citep{%
zoph2017learning,DBLP:conf/iclr/ZophL17} and genetic algorithms~\citep{%
DBLP:conf/aaai/RealAHL19,xie2017genetic} to search for transferable network blocks
whose performance surpasses many manually designed ones.  However, such
approaches require massive computation during the search --- typically thousands
of GPU days.  To reduce computational cost, recent efforts utilize more
efficient search techniques, such as direct gradient-based optimization~\citep{%
liu2018darts,
luo2018neural,
DBLP:conf/icml/PhamGZLD18,
tan2019mnasnet,
DBLP:conf/iclr/CaiZH19,
DBLP:conf/nips/WortsmanFR19}.
Nevertheless, most NAS methods perform search in a supernet space which
requires more computation than training typically-sized architectures.

\textbf{Network Growing.}
Network Morphism~\citep{DBLP:conf/icml/WeiWRC16} searches for efficient deep
networks by extending layers while preserving the parameters.  Recently
proposed Autogrow~\citep{DBLP:conf/kdd/Wen0CL20} takes an AutoML approach to
growing layers.  These methods either require a specially-crafted policy to
stop growth (\emph{e.g.,} after a fixed number of layers) or rely on evaluating
accuracy during training, incurring significant additional computational cost.

\textbf{Learning by Continuation.}
Continuation methods are commonly used to approximate intractable optimization
problems by gradually increasing the difficulty of the underlying objective,
for example by adopting gradual relaxations to binary problems.
\cite{DBLP:conf/cvpr/WuDZWSWTVJK19,
DBLP:conf/iclr/XieZLL19,
DBLP:conf/eccv/XieZZHL20} use
gumbel-softmax~\citep{DBLP:conf/iclr/JangGP17} to back-propagate errors during
architecture search and spatial feature sparsification.
\cite{DBLP:conf/nips/SavareseSM20} propose
\emph{continuous sparsification} to speed up pruning and ticket search~\citep{%
DBLP:conf/iclr/FrankleC19}.  Despite the success of continuation methods in
producing sparse networks upon the completion of training, they do not operate
on sparse networks during training and instead work with a real-valued
relaxation.  Postponing actual elimination of near zeroed-out components
prevents naive application of these methods from reducing training costs.

\section{Method}
\label{sec:method}

\subsection{Architectural Configuration Space}
\label{sec:grow_space}
A network topology can be seen as a directed acyclic graph consisting of an
ordered sequence of nodes.
Each node $\bx_{in}^{(i)}$ is an input feature and each edge is a computation
cell with \textit{structured} hyperparameters (\emph{e.g.,}~filter and layer
numbers in convolutional networks).
An architectural configuration space can be parameterized by associating a mask
variable $m \in \{0,1\}$ with each computation cell (edge), which enables
training-time pruning ($m=1\to0$) and growing ($m=0\to1$) dynamics.

As a running example, we consider a two-level configuration space for CNN
architectures, depicted in Figure~\ref{fig:growspace}, that enables dynamically
growing networks in both width (channel-wise) and depth (layer-wise).
Alternative configuration spaces are possible; we defer to the Appendix
details on how we parameterize the design of LSTM architectures.

\textbf{CNN Channel Configuration Space:}
For a convolutional layer with $l_{in}$ input channels, $l_{out}$ output
channels (filters) and $k \times k$ sized kernels, the $i$-th output feature is
computed based on the $i$-th filter,
\emph{i.e.}~for $i \in \{1, \dots, l_{out}\}$: 
\begin{eqnarray} \label{conv}
\bx_{out}^{(i)} = f(\bx_{in},\mathcal{F}^{(i)} \cdot \bm{m}_{c}^{(i)}) \,,
\end{eqnarray}
where $\bm{m}_{c}^{(i)} \in \{0,1\}$ is a binary parameter that removes the $i$-th
output channel when set to zero and $f$ denotes the convolutional operation.  $\bm{m}_c^{(i)}$ is shared across a filter and
broadcasts to the same shape as the filter tensor $\mathcal{F}^{(i)}$,
enabling growing/pruning of the entire filter.  As Figure~\ref{fig:growspace}
(top) shows, we start from a \emph{slim} channel configuration.  We then query
the indicator variables and perform \emph{state transitions}:
(1) When flipping an indicator variable from $0$ to $1$ for the first time, we
grow a randomly initialized filter and concatenate it to the network.
(2) If an indicator flips from $1$ to $0$, we temporarily detach the
corresponding filter from the computational graph; it will be grown back to the
its original position if its indicator flips back to $1$, or otherwise be
permanently pruned at the end of training.
(3) For other cases, the corresponding filters either survive and continue
training or remain detached pending the next query to their indicators.  Our
method automates architecture evolution, provided we can train the indicators.


\textbf{CNN Layer Configuration Space:}
To grow network depth, we design a layer configuration space in which an
initial shallow network will progressively expand into a deep trained model,
as shown in Figure~\ref{fig:growspace} (bottom).  Similar to channel
configuration space, where filters serve as basic structural units, we require
a unified formulation to support the growing of popular networks with shortcut
connections (\emph{e.g.,}~ResNets) and without (\emph{e.g.,}~VGG-like plain
nets).  We first introduce an abstract layer class $f_{layer}$ as a basic
structural unit, which operates on input features $\bx_{in}$ and generates
output features $\bx_{out}$.  $f_{layer}$ can be instantiated as convolutional
layers for plain nets or residual blocks for ResNets, respectively.  We define
the layer configuration space as:
\begin{eqnarray} \label{layerg}
\bx_{out} = g(\bx_{in}; f_{layer} \cdot \bm{m}_l^{(j)} ) = 
\begin{cases}
    f_{layer}(\bx_{in}), &  \text{if}  \quad \bm{m}_l^{(j)} = 1\\
    \bx_{in},              &  \text{if}  \quad \bm{m}_l^{(j)} = 0 
\end{cases}\,,
\end{eqnarray}
where $\bm{m}_l^{(j)} \in \{0,1\}$ is the binary indicator for $j$-th layer $f_{layer}$, with
which we perform state transitions analogous to the channel configuration
space.  Layer indicators have priority over channel indicators: if $\bm{m}_l^{(j)}$ is set
as 0, all filters contained in the corresponding layer will be detached,
regardless of the state their indicators.  We do not detach layers that perform
changes in resolution (\emph{e.g.,}~strided convolution).

\subsection{Growing with Structured Continuous Sparsification}
\label{sec:optim}
We can optimize a trade-off between accuracy and structured sparsity by
considering the objective:
\begin{eqnarray} \label{loss}
\min_{w, m_{c,l}, f_{layer}} \quad L_E(g(f(\bx; \bw \odot \bm{m}_{c}); f_{layer} \cdot \bm{m}_l)) + \lambda_1 \left\lVert \bm{m}_c \right\rVert_0 + \lambda_2 \left\lVert \bm{m}_l \right\rVert_0
\,,
\end{eqnarray}
where $f$ is the operation in Eq.~(\ref{conv}) or Eq.~(\ref{LSTM})
(in Appendix~\ref{lstm_ext}), while $g$ is defined in Eq.~(\ref{layerg}).
$\bw \odot \bm{m}_c$ and $f_{layer} \cdot \bm{m}_l$ are general expressions of structured
sparsified filters and layers and $L_E$ denotes a loss function
(\emph{e.g.,}~cross-entropy loss for classification).  The $\ell_0$ terms
encourage sparsity, while $\lambda_{1,2}$ are trade-off parameters between $L_E$
and the $\ell_0$ penalties.

\begin{wrapfigure}{R}{0.5\textwidth}
\begin{minipage}{0.5\textwidth}
\vspace{-2.5em}
\begin{algorithm}[H]
\caption{: Optimization}
\label{alg:optimization}
\begin{algorithmic}
   \STATE {\bfseries Input:} Data $\bm{X}$ = $( \bm{x}_i)_{i=1}^{n}$, labels $\bm{Y}$ = $( \bm{y}_i)_{i=1}^{n}$
   \STATE {\bfseries Output:} Grown efficient model $G$
   \STATE Initialize: $G$, $w$, $u$, $\lambda_1^{\text{base}}$ and $\lambda_2^{\text{base}}$.
   \STATE Set $\bt_s$ as all 0 vectors associating $\sigma$ functions.
   \FOR{$\text{epoch}=1$ {\bfseries to} $T$}
      \STATE Evaluate $G$'s sparsity $u_G$ and calculate\\
         \quad $\Delta u = u - u_G$
      \STATE Update
         $\lambda_1 \leftarrow \lambda_1^{\text{base}} \cdot \Delta u$;
         $\lambda_2 \leftarrow \lambda_2^{\text{base}} \cdot \Delta u$\\
         \quad in Eq.~(\ref{final_loss}) using Eq.~(\ref{eq:update_lambda})
      \FOR{$r=1$ {\bfseries to} $R$}
         \STATE Sample mini-batch $x_i, y_i$ from $\bX,\bY$
         \STATE Train $G$ using Eq.~(\ref{final_loss}) with SGD
      \ENDFOR
      \STATE Sample indicators $q_{c,l} \sim \text{Bern}(\sigma(\beta s_{c,l} ))$
      \STATE and record the index $idx$ where $q$ value is 1.
      \STATE Update $\bt_s[idx] = \bt_s[idx] + 1$
      \STATE Update $\bm{\beta}$ using Eq.~(\ref{temp_scheduler_local})
   \ENDFOR
   \STATE return G
\end{algorithmic}
\end{algorithm}
\end{minipage}
\end{wrapfigure}

\textbf{Budget-aware Growing.}
In practice, utilizing Eq.~(\ref{loss}) might require a grid search on $\lambda_1$
and $\lambda_2$ until a network with desired sparsity is produced.  To avoid such
a costly procedure, we propose a budget-aware growing process, guided by a target
budget in terms of model parameters or FLOPs.  Instead of treating $\lambda_1$ and
$\lambda_2$ as constants, we periodically update them as:
\begin{eqnarray} \label{eq:update_lambda}
\lambda_1 \leftarrow \lambda_1^{\text{base}} \cdot \Delta u, \lambda_2 \leftarrow \lambda_2^{\text{base}} \cdot \Delta u \,,
\end{eqnarray}
where $\Delta u$ is calculated as the target sparsity $u$ minus current network
sparsity $u_G$, and $\lambda_1^{\text{base}}$, $\lambda_2^{\text{base}}$ are
initial base constants.  In early growing stages, since the network is too sparse
and $\Delta u$ is negative, the optimizer will drive the network towards a state
with more capacity (wider/deeper).  The regularization effect gradually weakens as
the network's sparsity approaches the budget (and $\Delta u$ approaches zero).
This allows us to adaptively grow the network and automatically adjust its
sparsity level while simultaneously training model weights.
Appendix~\ref{append:buget_grow} provides more detailed analysis.  Our
experiments default to defining budget by parameter count, but also investigate
alternative notions of budget.

\textbf{Learning by Continuation.}
Another issue in optimizing Eq.~(\ref{loss}) is that $\left\lVert m_c \right\rVert_0$ and $\left\lVert m_l \right\rVert_0$ make the problem computationally intractable due to the combinatorial nature of binary states.
To make the configuration space continuous and the optimization feasible, we borrow the concept of learning by continuation~\citep{DBLP:conf/iccv/CaoLWY17, DBLP:conf/cvpr/WuDZWSWTVJK19, DBLP:conf/nips/SavareseSM20,DBLP:conf/eccv/XieZZHL20}.
We reparameterize $m$ as the binary sign of a continuous variable $s$: $\sign(s)$ is 1 if $s > 0$ and 0 if $s < 0$. 
We rewrite the objective in Eq.~(\ref{loss}) as:
\begin{eqnarray} \label{sign_loss}
\min_{\bw, \bs_{c,l}\neq0, f_{layer}}  L_E \Big( g \big( f(\bx; \bw \odot \sign(\bs_c) ); f_{layer} \cdot \sign(\bs_l) \big) \Big) + \lambda_1 \left\lVert \sign(\bs_c)  \right\rVert_1 + \lambda_2 \left\lVert \sign(\bs_l)  \right\rVert_1 \,.
\end{eqnarray}
We attack the hard and discontinuous optimization problem in Eq.~(\ref{sign_loss}) by starting with an \textit{easier} objective which becomes \textit{harder} as training proceeds. 
We use a sequence of functions whose limit is the sign operation: for any $s \neq 0$,
$\lim_{\beta \to \infty}\sigma(\beta s) = \sign(s)$ if $\sigma$ is sigmoid function or $\lim_{\beta \to 0}\sigma(\beta s) = \sign(s)$ if $\sigma$ is gumbel-softmax $\frac{exp((-log(s_0) + g_1(s))/\beta)}{\sum_{j \in \{0,1\}}exp((-log(s_j) + g_j(s))/\beta)}$~\citep{DBLP:conf/iclr/JangGP17}, 
where $\beta > 0$ is a temperature parameter and $g_{0,1}$ is gumbel. By periodically changing $\beta$, $\sigma(\beta s)$ becomes harder to optimize, while the objectives converges to original discrete one. 

\textbf{Maintaining Any-time Sparsification.}
Although continuation methods can make the optimization feasible, they only conduct sparsification via a thresholding criterion in the inference phase. In this case, the train-time architecture is dense and not appropriate in the context of growing a network. 
To effectively reduce computational cost of training, we maintain a sparse architecture by introducing an 0-1 sampled auxiliary variable $q$ based on the probability value $\sigma(\beta s)$.
Our final objective becomes:
\begin{eqnarray} \label{final_loss}
\min_{\bw, \bs_{c,l}\neq0, f_{layer}} L_E \Big( g \big(f(\bx; \bw \odot \sigma(\beta \bs_c) \odot \bq_c); f_{layer} \cdot \sigma(\beta \bs_l) \cdot \bq_l \big) \Big) + \lambda_1 \left\lVert \sigma(\beta \bs_c)  \right\rVert_1 + \lambda_2 \left\lVert \sigma(\beta \bs_l)  \right\rVert_1 
\,,
\end{eqnarray}
where $\bq_c$ and $\bq_l$ are random variables sampled from $\text{Bern}(\sigma(\beta \bs_c))$ and $\text{Bern}(\sigma(\beta \bs_l))$, which effectively maintains any-time sparsification and avoids sub-optimal thresholding, as shown in Figure~\ref{fig:growprocedure}.

\textbf{Improved Temperature Scheduler.}
In existing continuation methods, the initial $\beta$ value is usually set as $\beta_0=1$ and a scheduler is used at the end of each training epoch to update $\beta$ in all activation functions $\sigma$, typically following $\beta = \beta_0 \cdot \gamma^t$,
where $t$ is current epoch and $\gamma$ is a hyperparameter ($>1$ when $\sigma$ is the sigmoid function, $<1$ when $\sigma$ is gumbel softmax). Both $\gamma$ and $t$ control the speed at which the temperature increases during training. Continuation methods with global temperature schedulers have been successfully applied
in pruning and NAS.  However, in our case, a global schedule leads to unbalanced dynamics between variables with low and high sampling probabilities: increasing the temperature of those less sampled at early stages may hinder their training altogether, as towards the end of training the optimization difficulty is higher.
To overcome this issue, we propose a separate, structure-wise temperature scheduler by making a simple modification: for each mask variable, instead of using the current epoch number $t$ to compute its temperature, we set a separate counter
$\bt_s$ which is increased only when its associated indicator variable is sampled as 1 in Eq.~(\ref{final_loss}). We define our structure-wise temperature scheduler as 
\begin{eqnarray} \label{temp_scheduler_local}
\bm{\beta} = \beta_0  \cdot \gamma^{\bt_s} \,,
\end{eqnarray}
where $\bt_s$ are vectors associated with the $\sigma$ functions.
Experiments use this separate scheduler by default, but also compare the two alternatives.
Algorithm~\ref{alg:optimization} summarizes our optimization procedure.

\section{Experiments}
\label{sec:experiments}

We evaluate our method against existing channel pruning, network growing, and
neural architecture search (NAS) methods on:
CIFAR-10~\citep{krizhevsky2014cifar} and ImageNet~\citep{deng2009imagenet} for
image classification,
PASCAL~\citep{DBLP:journals/ijcv/EveringhamEGWWZ15} for semantic segmentation
and the Penn Treebank (PTB)~\citep{DBLP:journals/coling/MarcusSM94} for
language modeling.  See dataset details in Appendix~\ref{append_dataset}.
In tables, best results are highlighted in bold and second best are underlined.

\begin{table}[tb]
\vspace{-3em}
\begin{center}
\setlength{\tabcolsep}{4pt}
\footnotesize
\caption{\footnotesize{Comparison with the channel pruning methods L1-Pruning~\citep{DBLP:conf/iclr/0022KDSG17}, SoftNet~\citep{DBLP:conf/ijcai/HeKDFY18}, ThiNet~\citep{DBLP:conf/iccv/LuoWL17}, Provable~\citep{DBLP:conf/iclr/LiebenweinBLFR20} and BAR~\citep{DBLP:conf/cvpr/LemaireAJ19} on CIFAR-10.}}
\label{tab:cifar10:result}
\vspace{-5pt}
\begin{tabular}{@{}cccccc@{}}
\toprule
{Model} & {Method} & {Val Acc(\%)} & {Params(M)} & {FLOPs(\%)} & {Train-Cost Savings($\times$)}\\
\midrule
~ &Original &92.9 $\pm$ 0.16 (-0.0) &14.99 (100\%) &100  &1.0$\times$\\
~ &L1-Pruning &91.8 $\pm$ 0.12 (-1.1) &2.98 (19.9\%) &19.9 &2.5$\times$\\
VGG &SoftNet  &92.1 $\pm$ 0.09 (-0.8) &5.40 (36.0\%) &36.1 &1.6$\times$\\
-16 &ThiNet &90.8 $\pm$ 0.11 (-2.1) &5.40 (36.0\%) &36.1 &1.6$\times$\\
~ &Provable &\underline{92.4 $\pm$ 0.12 (-0.5)} &\underline{0.85 (5.7\%)} &\underline{15.0} &\underline{3.5}$\times$\\
~ &Ours
  &\textbf{92.50 $\pm$ 0.10 (-0.4)} &\textbf{0.754 $\pm$ 0.005 (5.0\%)} &\textbf{13.55 $\pm$ 0.03} &\textbf{4.95 $\pm$ 0.17 $\times$}\\
\midrule
~ &Original &91.3 $\pm$ 0.12 (-0.0) &0.27 (100\%) &100  &1.0$\times$\\
~ &L1-Pruning &\underline{90.9 $\pm$ 0.10} (-0.4) &0.15 (55.6\%) &55.4 &1.1$\times$ \\
ResNet &SoftNet &90.8 $\pm$ 0.13 (-0.5) &0.14 (53.6\%) &\underline{50.6} &1.2$\times$\\
-20 &ThiNet &89.2 $\pm$ 0.18 (-2.1) &0.18 (67.1\%) &67.3 &1.1$\times$\\
~ &Provable &90.8 $\pm$ 0.08 (-0.5) &\underline{0.10 (37.3\%)} &54.5 &\underline{1.7}$\times$\\
~ &Ours &\textbf{90.91 $\pm$ 0.07 (-0.4)} &\textbf{0.096 $\pm$ 0.002 (35.8\%)} &\textbf{50.20 $\pm$ 0.01} &\textbf{2.40 $\pm$ 0.09 }$\times$\\
\midrule
\multirow{2}{*}{WRN} &Original &96.2 $\pm$ 0.10 (-0.0) &36.5 (100\%) &100 &1.0$\times$\\
\multirow{2}{*}{-28} &L1-Pruning &\underline{95.2 $\pm$ 0.10 (-1.0)} &7.6 (20.8\%) &49.5 &1.5$\times$ \\
\multirow{2}{*}{-10} &BAR(16x V) &92.0 $\pm$ 0.08 (-4.2) &\textbf{2.3 (6.3\%)} &\textbf{1.5} &\underline{2.6}$\times$\\
~ &Ours &\textbf{95.32 $\pm$ 0.11 (-0.9)} &\underline{3.443 $\pm$ 0.010 (9.3\%)} &\underline{28.25 $\pm$ 0.04} &\textbf{3.12  $\pm$ 0.11}$\times$\\
\bottomrule
\end{tabular}
\end{center}
\vspace{-1em}
\end{table}

\vspace{-1.0em}
\subsection{Comparing with Channel Pruning Methods}
\vspace{-0.5em}
\textbf{Implementation Details.}
For fair comparison, we only grow filters while keeping other structured
parameters of the network (number of layers/blocks) the same as unpruned
baseline models.  Our method involves two kinds of trainable variables: model
weights and mask weights. For model weights, we adopt the same hyperparameters
used to train the corresponding unpruned baseline models, except for setting
the dropout keep probability for language modeling to 0.65.  We initialize mask
weights such that a single filter is activated in each layer.  We train with
SGD, an initial learning rate of 0.1, weight decay of $10^{-6}$ and momentum
0.9. Trade-off parameter $\lambda_1^{\text{base}}$ is set to 0.5 on all tasks;
$\lambda_2$ is not used since we do not perform layer growing here.
We set $\sigma$ as the sigmoid function and $\gamma$ as $100^{\frac{1}{T}}$
where $T$ is the total number of epochs.

\vspace{-0.25em}
\textbf{VGG-16, ResNet-20, and WideResNet-28-10 on CIFAR-10.}
Table~\ref{tab:cifar10:result} summarizes the models produced by our method
and competing channel pruning approaches.
Note that training cost is
calculated based on overall FLOPs during pruning and growing stages.
Our method produces sparser networks with less accuracy degradation, and,
consistently saves more computation during training --- a consequence of
growing from a simple network.
For a aggressively pruned WideResNet-28-10, we observe that
BAR~\citep{DBLP:conf/cvpr/LemaireAJ19} might not have enough capacity to
achieve negligible accuracy drop, even with knowledge
distillation~\citep{DBLP:journals/corr/HintonVD15} during training.
Note that we report our method's performance as mean $\pm$ standard deviation, computed over 5 runs with different random seeds. The small observed variance shows that our method performs consistently across runs.

\vspace{-0.25em}
\textbf{ResNet-50 and MobileNetV1 on ImageNet.}
To validate effectiveness on large-scale datasets, we grow, from scratch,
filters of the widely used ResNet-50 on ImageNet; we do not fine-tune.
Table~\ref{tab:imagenet:result} shows our results best those directly reported
in papers of respective competing methods.  Our approach achieves $49.7\%$
inference and $47.4\%$ training cost savings in terms of FLOPs while
maintaining $75.2\%$ top-1 accuracy, without any fine-tuning stage.
Appendix~\ref{mobile_imagenet} shows our improvements on the challenging
task of growing channels of an already compact MobileNetV1.  In addition, Figure~\ref{fig:netadapt_tradeoff} shows the top-1 accuracy/FlOPs trade-offs for MobileNetV1 on ImageNet, demonstrating that our method dominates competing approaches.

\vspace{-0.25em}
\textbf{Deeplab-v3-ResNet-101 on PASCAL VOC.}
Appendix~\ref{deeplab} provides semantic segmentation results.

\vspace{-0.25em}
\textbf{2-Stacked-LSTMs on PTB:}
We detail the extensions to recurrent cells and compare our proposed method
with ISS based on vanilla two-layer stacked LSTM in Appendix~\ref{lstm_ext}.
As shown in Table~\ref{tab:ptb:result}, our method finds more compact model structure with lower training cost, while achieving similar perplexity on both validation and test sets.

\begin{table}[tb]
\vspace{-3em}
\begin{center}
\setlength{\tabcolsep}{10.75pt}
\footnotesize
\caption{\footnotesize{Comparison with channel pruning methods: L1-Pruning~\citep{DBLP:conf/iclr/0022KDSG17}, SoftNet~\citep{DBLP:conf/ijcai/HeKDFY18} and Provable~\citep{DBLP:conf/iclr/LiebenweinBLFR20} on ImageNet.}}
\label{tab:imagenet:result}
\vspace{1pt}
\begin{tabular}{@{}cccccc@{}}
\toprule
{Model} & {Method} & {Top-1 Acc(\%)} & {Params(M)} & {FLOPs(\%)} & {Train-Cost Savings($\times$)}\\
\midrule
~ &Original &76.1 (-0.0) &23.0 (100\%) &100 & 1.0($\times$)\\
\multirow{2}{*}{ResNet} &L1-Pruning &74.7 (-1.4) &19.6 (85.2\%) &77.5 & 1.1($\times$)\\
\multirow{2}{*}{-50} &SoftNet &74.6 (-1.5) &N/A &\underline{58.2} & \underline{1.2}($\times$)\\
~ &Provable &\textbf{75.2 (-0.9)} &\underline{15.2 (65.9\%)} &70.0 & \underline{1.2}($\times$)\\
~ &Ours &\textbf{75.2 (-0.9)} &\textbf{14.1 (61.2\%)} &\textbf{50.3} & \textbf{1.9}($\times$)\\
\bottomrule
\end{tabular}
\end{center}
\end{table}
\begin{table}[tb]
\vspace{-2em}
\begin{center}
\setlength{\tabcolsep}{7.75pt}
\footnotesize
\caption{\footnotesize{Results comparing with AutoGrow~\citep{DBLP:conf/kdd/Wen0CL20} on CIFAR-10 and ImageNet.}}
\vspace{1pt}
\label{tab:autogrow:result}
\begin{tabular}{@{}ccccccc@{}}
\toprule
{Dataset} & {Methods} & {Variants} & {Found Net} & {Val Acc(\%)} & {Depth} & {Sparse Channel}\\
\cline{1-7}
~ & \multirow{2}{*}{Ours} &\textit{Basic3ResNet} &23-29-31 &\textbf{94.50} &\textbf{83} &\cmark\\
\multirow{2}{*}{CIFAR-10} & &\textit{Plain3Net} &11-14-19 &\textbf{90.99} &\textbf{44} &\cmark\\
\cline{2-7}
~  & \multirow{2}{*}{AutoGrow} &\textit{Basic3ResNet} &42-42-42 &94.27 &126 &\xmark \\
~ & &\textit{Plain3Net}  &23-22-22 &90.82 &67 &\xmark\\
\cline{1-7}
~  & \multirow{2}{*}{Ours} &\textit{Bottleneck4ResNet} &5-6-5-7 &\textbf{77.41} &\textbf{23} &\cmark\\
\multirow{2}{*}{ImageNet} & &\textit{Plain4Net} &3-4-4-5 &\textbf{70.79} &\textbf{16} &\cmark\\
\cline{2-7}
~  & \multirow{2}{*}{AutoGrow} &\textit{Bottleneck4ResNet} &6-7-3-9 &77.33 &25 &\xmark \\
~ & &\textit{Plain4Net} &5-5-5-4 &70.54 &19  &\xmark\\
\bottomrule
\end{tabular}
\end{center}
\vspace{-2em}
\end{table}

\vspace{-0.5em}
\subsection{Comparing with AutoGrow}

\vspace{-0.5em}
\textbf{Implementation Details.}
We grow both filters and layers.  We follow AutoGrow's settings in exploring
architectural variations that define our initial seed network, layer-wise
configuration space and basic structural units $f_{layer}$:
\textit{Basic3ResNet},
\textit{Bottleneck4ResNet},
\textit{Plain3Net}, \textit{Plain4Net}.
Different from the initialization of AutoGrow using full-sized filters in each
layer, our channel-wise configuration space starts from single filter and
expands simultaneously with layers.  Appendix~\ref{seed_arch} contains a
detailed comparison of seed architectures.  For training model weights, we
adopt the hyperparameters of the ResNet or VGG models corresponding to initial
seed variants.  For layer-wise and channel-wise mask variables, we initialize
the weights such that only a single filter in each layer and one basic unit in
each stage (\emph{e.g.,} BasicBlock in \textit{Basic3ResNet}) is active.
We use SGD training with initial learning rate of 0.1, weight decay of
$10^{-6}$ and momentum of 0.9 on all datasets. The learning rate scheduler
is the same as for the corresponding model weights.  Trade-off parameters
$\lambda_1^{\text{base}}$ and $\lambda_2^{\text{base}}$ are set as 1.0 and 0.1
on all datasets.  For fair comparison, we fine-tune our final models with
40 epochs and 20 epochs on CIFAR-10 and ImageNet, respectively.

\vspace{-0.25em}
\textbf{Results on CIFAR-10 and ImageNet.}
Table~\ref{tab:autogrow:result} compares our results to those of AutoGrow.  For
all layer-wise growing variants across both datasets, our method consistently
yields a better depth and width configuration than AutoGrow, in terms of
accuracy and training/inference costs trade-off.  Regarding the training time
of \textit{Bottleneck4ResNet} on ImageNet, AutoGrow requires $61.6$ hours for
the growing phase and $78.6$ hours for fine-tuning on 4 TITAN V GPUs, while
our method takes $48.2$ and $31.3$ hours, respectively.  Our method offers
$43\%$ more train-time savings than AutoGrow.  We not only require fewer
training epochs, but also grow from a single filter to a relatively sparse
network, while AutoGrow always keeps full-sized filter sets without any
reallocation.

\vspace{-0.5em}
\subsection{Comparing with NAS Methods}

\vspace{-0.5em}
As a fair comparison with NAS methods involving search and re-training phases,
we also divide our method into growing and training phases.
Specifically, we grow layers and channels from the \textit{Bottleneck4ResNet}
seed architecture directly on ImageNet by setting $\lambda_1^{base}=2.0$,
$\lambda_2^{base}=0.1$ and the parameter budget under 7M.
Then we resume training the grown architecture and compare with existing NAS
methods in terms of parameters, top-1 validation accuracy and V100 GPU hours
required by the search or growing stages, as shown in Table~\ref{tab:nas}.
Note that DARTS~\citep{liu2018darts} conducts search on
CIFAR-10, then transfers to ImageNet instead of direct search.  This is because
DARTS operates on a supernet by including all the candidate paths and suffers
from GPU memory explosion.  In terms of epoch-wise FLOPs, results shown in
Figure~\ref{fig:motivation}(c) are for training an equivalent of ResNet-20 on
CIFAR-10 in comparison with DARTS and Provable channel pruning
approach~\citep{DBLP:conf/iclr/LiebenweinBLFR20}.
Also note that the EfficientNet-B0 architecture, included in Table~\ref{tab:nas}, is generated by grid search in the MnasNet search space, thus having the same heavy search cost. To achieve the reported performance, EfficientNet-B0 utilizes additional squeeze-and-excitation (SE)~\citep{DBLP:conf/cvpr/HuSS18} modules, AutoAugment~\citep{DBLP:journals/corr/abs-1805-09501}, as well as much longer re-training epochs on ImageNet.
\begin{wraptable}{r}{8.5cm}
\vspace{-1.0em}
\setlength{\tabcolsep}{5pt}
\footnotesize
\caption{\footnotesize{Performance comparing with NAS methods AmoebaNet-A~\citep{DBLP:conf/aaai/RealAHL19}, MnasNet~\citep{tan2019mnasnet}, EfficientNet-B0~\citep{DBLP:conf/icml/TanL19}, DARTS~\citep{liu2018darts} and ProxylessNet~\citep{DBLP:conf/iclr/CaiZH19} on ImageNet.}}\label{tab:nas}
\begin{tabular}{@{}cccc@{}}\\\toprule
{Method} & {Params} & {Top-1} & {Search/Grow Cost}\\
\midrule
AmoebaNet-A &5.1M &74.5\% &76K GPU hours\\
MnasNet &\textbf{4.4M} &74.0\% &40K GPU hours\\
EfficientNet-B0 &5.3M &\textbf{77.1}\% (+SE) &40K GPU hours\\
DARTS &\underline{4.7M} &73.1\% &N/A \\
\midrule
ProxylessNet(GPU) &7.1M &\underline{75.1\%} &200 GPU hours \\
Ours &6.8M &74.3\% &\textbf{80} GPU hours\\
Ours &6.7M &74.8\% &\underline{110} GPU hours\\
Ours &6.9M &\underline{75.1\%} &140 GPU hours\\
\bottomrule
\end{tabular}
\vspace{-1.5em}
\end{wraptable}

ProxylessNet still starts with an over-parameterized supernet, but applies a
pruning-like search method by binarizing the architecture parameters and
forcing only one path to be activated at search-time.  This
enables directly searching on ImageNet, achieving $200 \times$ more search
cost savings than MnasNet.  Contrasting with ProxylessNet, our method
progressively adds filters and layers to simple seed architectures while
maintaining sparsification, which leads to savings of not only epoch-wise
computation but also memory consumption, enabling faster, larger-batch
training.  As such, we further save $45\%$ of the GPU search hours, while
achieving comparable accuracy-parameter trade-offs.

\vspace{-1.0em}
\subsection{Analysis}
\begin{figure}[tb]
   \vspace{-1 em}
   \begin{minipage}[t]{1.0\linewidth}
   \begin{minipage}[t]{0.42\linewidth}
      \centering
      \vspace{0pt}
      \includegraphics[width=\linewidth]{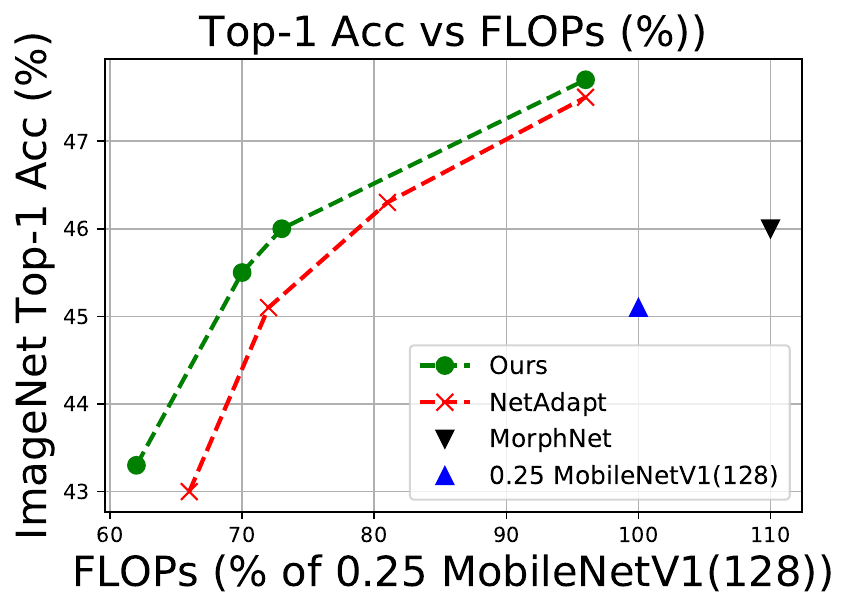}
      \vspace{-20pt}
      \caption{\footnotesize{Performance/FLOPs trade-offs for pruned MobileNetV1 on ImageNet.}}
      \label{fig:netadapt_tradeoff}
   \end{minipage}
   \hfill
   \begin{minipage}[t]{0.55\linewidth}
      \centering
      \vspace{2.5pt}
      \includegraphics[width=\linewidth]{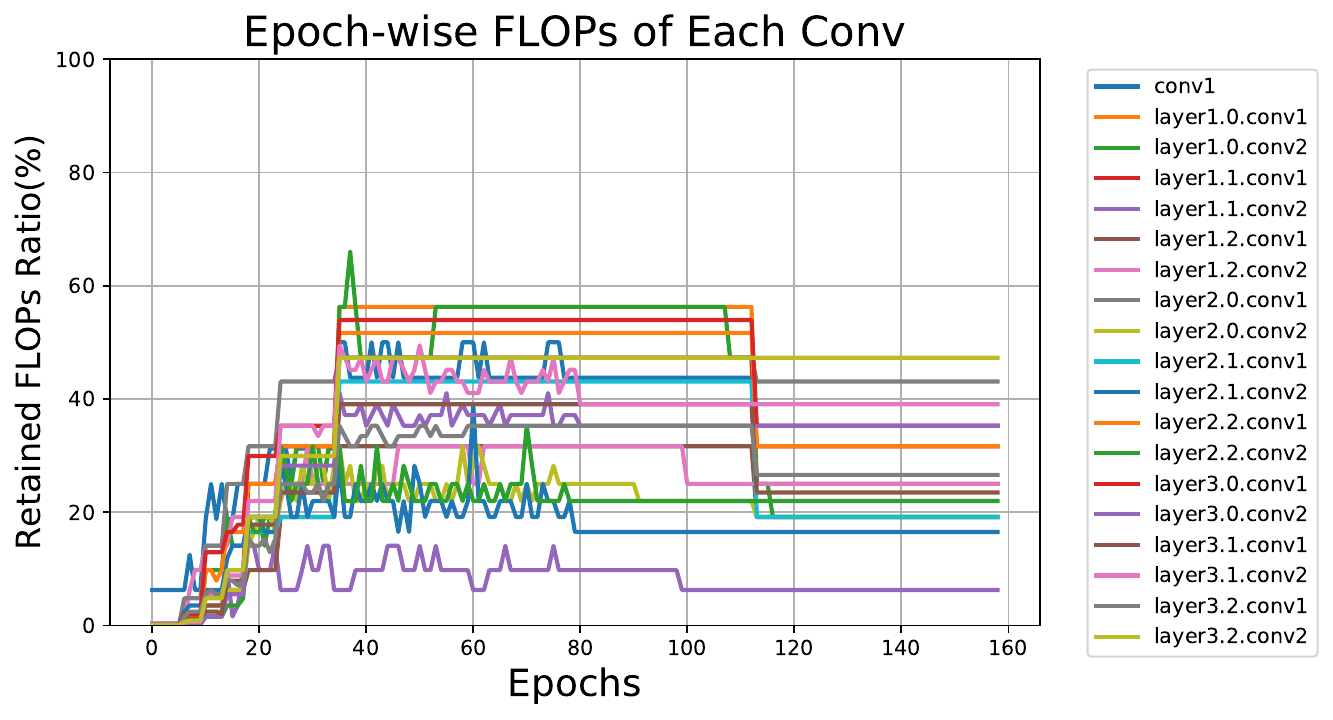}
      \vspace{-20pt}
      \caption{\footnotesize{Epoch-wise training FLOPs for channel growing a ResNet-20.}}
      \label{fig:res20_track}
   \end{minipage}
   \end{minipage}
   ~\vspace{-5pt}\\
   \begin{minipage}[t]{1.0\linewidth}
   \begin{minipage}[t]{0.42\linewidth}
      \centering
      \vspace{0pt}
      \includegraphics[width=\linewidth]{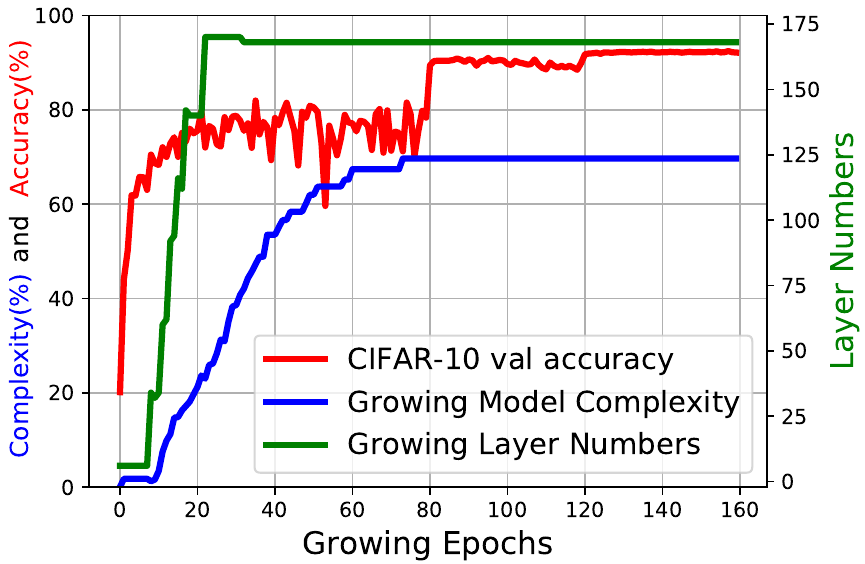}
      \vspace{-20pt}
      \caption{\footnotesize{Tracking validation accuracy, complexity and layers for \textit{Basic3ResNet} growing.}}
      \label{fig:b3res_track}
   \end{minipage}
   \hfill
   \begin{minipage}[t]{0.55\linewidth}
      \centering
      \vspace{-4pt}
      \includegraphics[width=0.70\linewidth]{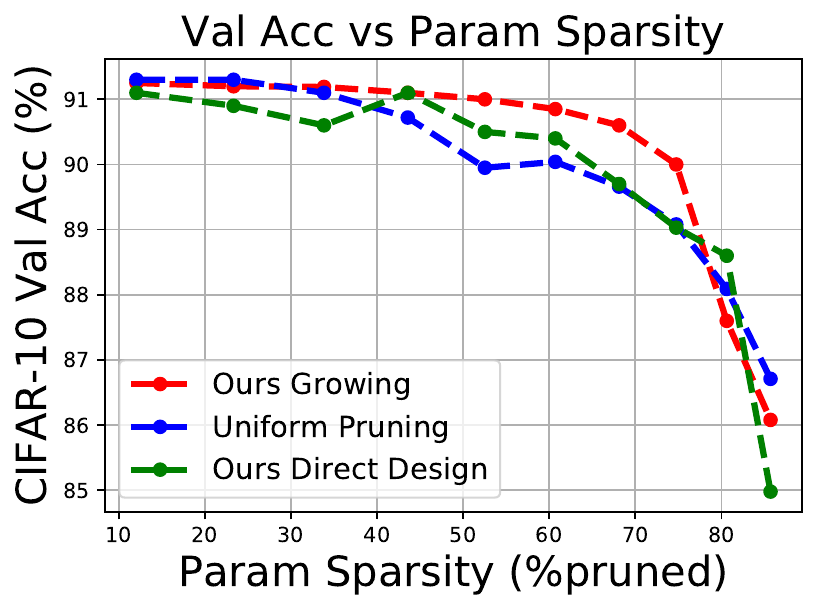}
      \vspace{-10pt}
      \caption{\footnotesize{Pruned architectures obtained by ablated methods with different parameter sparsity.}}
      \label{fig:acc_sparse}
   \end{minipage}
   \end{minipage}
   \vspace{-1em}
\end{figure}

\vspace{-0.5em}
\textbf{Training Cost Savings.}
Figure~\ref{fig:res20_track} illustrates our sparsification dynamics,
showing epoch-wise FLOPs while growing a ResNet-20.
Appendix~\ref{sec:append_anytime_sparse} presents additional visualizations.
Even with fully parallel GPU hardware, starting with few filters and layers in
the network will ultimately save wall-clock time, as larger batch
training~\citep{DBLP:journals/corr/GoyalDGNWKTJH17} can always be employed to
fill the hardware.

Figure~\ref{fig:b3res_track} shows validation accuracy, model complexity, and
layer count while growing \textit{Basic3ResNet}. Complexity is measured as the
model parameters ratio of AutoGrow's target model.  At the end of 160 epochs,
our method's validation accuracy is $92.36\%$ , which is higher than
AutoGrow's $84.65\%$ at 360 epochs. We thus require fewer fine-tuning
epochs to achieve a final $94.50\%$ accuracy on CIFAR.
\begin{wraptable}{r}{8.5cm}
\setlength{\tabcolsep}{10.75pt}
\footnotesize
\caption{\footnotesize{Comparison with random pruning baseline on CIFAR-10.}}\label{tab:random}
\vspace{-5pt}
\begin{tabular}{@{}cccc@{}}\\\toprule
{Model} & {Method} & {Val Acc(\%)} & {Params(M)}\\
\midrule
\multirow{2}{*}{VGG-16} & Random &90.01 $\pm$ 0.69 & 0.770 $\pm$ 0.050 \\
~ &Ours &\textbf{92.50 $\pm$ 0.10} &0.754 $\pm$ 0.005 \\
\midrule
\multirow{2}{*}{ResNet-20} & Random &89.18 $\pm$ 0.55 &0.100 $\pm$ 0.010 \\
~ &Ours &\textbf{90.91 $\pm$ 0.07}   &0.096 $\pm$ 0.002 \\
\midrule
\multirow{2}{*}{WRN-28-10} & Random &92.26 $\pm$ 0.87 &3.440 $\pm$ 0.110\\
~ &Ours &\textbf{95.32 $\pm$ 0.11} &3.443 $\pm$ 0.010\\
\bottomrule
\end{tabular}
\end{wraptable}

\textbf{Budget-Aware Growing.}
In Figure~\ref{fig:acc_sparse}, for ResNet-20 on CIFAR-10, we compare
architectures obtained by (1) \textit{uniform pruning}: a naive predefined
pruning method that prunes the same percentage of channels in each layer,
(2) \textit{ours}: variants of our method by setting different model parameter
sparsities as target budgets during growing, and
(3) \textit{direct design}: our grown architectures re-initialized with random
weights and re-trained.  In most budget settings, our growing method
outperforms direct design and uniform pruning, demonstrating higher parameter
efficiency.  Our method also appears to have positive effect in terms of
regularization or optimization dynamics, which are lost if one attempts to
directly train the final compact structure.
Appendix~\ref{flops_budget_append} investigates FLOPs-based budget targets.

\textbf{Comparing with Random Baseline.}
In addition to the \textit{uniform pruning} baseline in Figure~\ref{fig:acc_sparse}, we also compare with a random sampling baseline to further separate the contribution of our configuration space and growing method, following the criterion in~\citep{
DBLP:conf/iccv/XieKGH19,
DBLP:conf/uai/LiT19,
DBLP:conf/iclr/YuSJMS20,
DBLP:conf/iccv/Radosavovic0XLD19}.
Specifically, this random baseline replaces the procedure for sampling entries of $\bm{q}$ in Eq.~\ref{final_loss}.  Instead of using sampling probabilities derived from the learned mask parameters $\bm{s}$, it samples with fixed probability.  As shown in Table~\ref{tab:random}, our method consistently performs much better than this random baseline.  These results, as well as the more sophisticated baselines in Figure~\ref{fig:acc_sparse}, demonstrate the effectiveness of our growing and pruning approach.

\textbf{Temperature Scheduler.}
We compare our structure-wise temperature control to a global one in channel
growing experiments on CIFAR-10 using VGG-16, ResNet-20, and WideResNet-28-10.
Table~\ref{tab:cifar10:result} results use our structure-wise scheduler.
To achieve similar sparsity with the global scheduler, the corresponding models
suffer accuracy drops of $1.4\%$, $0.6\%$, and $2.5\%$.  With the global
scheduler, optimization of mask variables stops early in training and the
following epochs are equivalent to directly training a fixed compact network.
This may force the network to be stuck with a suboptimal architecture.
Appendix~\ref{interaction_lr_temp} investigates learning rate and temperature
schedule interactions.

\section{Conclusion}
\label{sec:conclusion}

We propose a simple yet effective method to grow efficient deep networks via structured continuous sparsification, which decreases the computational cost not only of inference but also of training. The method is simple to implement and quick to execute; it automates the network structure reallocation process under practical resource budgets. Application to widely used deep networks on a variety of tasks shows that our method consistently generates models with better accuracy-efficiency trade-offs than competing methods, while achieving considerable training cost savings.

~\\~\\
\textbf{Acknowledgments.}
This work was supported by the University of Chicago CERES Center for
Unstoppable Computing and the National Science Foundation under grant
CNS-1956180.

\bibliography{iclr2021_conference}

\begin{thebibliography}{73}
\providecommand{\natexlab}[1]{#1}
\providecommand{\url}[1]{\texttt{#1}}
\expandafter\ifx\csname urlstyle\endcsname\relax
  \providecommand{\doi}[1]{doi: #1}\else
  \providecommand{\doi}{doi: \begingroup \urlstyle{rm}\Url}\fi

\bibitem[Cai et~al.(2019)Cai, Zhu, and Han]{DBLP:conf/iclr/CaiZH19}
Han Cai, Ligeng Zhu, and Song Han.
\newblock Proxylessnas: Direct neural architecture search on target task and
  hardware.
\newblock In \emph{ICLR}, 2019.

\bibitem[Cao et~al.(2017)Cao, Long, Wang, and Yu]{DBLP:conf/iccv/CaoLWY17}
Zhangjie Cao, Mingsheng Long, Jianmin Wang, and Philip~S. Yu.
\newblock Hashnet: Deep learning to hash by continuation.
\newblock In \emph{ICCV}, 2017.

\bibitem[Chen et~al.(2017)Chen, Papandreou, Schroff, and
  Adam]{DBLP:journals/corr/ChenPSA17}
Liang{-}Chieh Chen, George Papandreou, Florian Schroff, and Hartwig Adam.
\newblock Rethinking atrous convolution for semantic image segmentation.
\newblock \emph{arXiv:1706.05587}, 2017.

\bibitem[Cho et~al.(2014)Cho, van Merrienboer, G{\"{u}}l{\c{c}}ehre, Bahdanau,
  Bougares, Schwenk, and Bengio]{DBLP:conf/emnlp/ChoMGBBSB14}
Kyunghyun Cho, Bart van Merrienboer, {\c{C}}aglar G{\"{u}}l{\c{c}}ehre, Dzmitry
  Bahdanau, Fethi Bougares, Holger Schwenk, and Yoshua Bengio.
\newblock Learning phrase representations using {RNN} encoder-decoder for
  statistical machine translation.
\newblock In \emph{EMNLP}, 2014.

\bibitem[Chollet(2017)]{DBLP:conf/cvpr/Chollet17}
Fran{\c{c}}ois Chollet.
\newblock Xception: Deep learning with depthwise separable convolutions.
\newblock In \emph{CVPR}, 2017.

\bibitem[Cubuk et~al.(2019)Cubuk, Zoph, Man{\'{e}}, Vasudevan, and
  Le]{DBLP:journals/corr/abs-1805-09501}
Ekin~Dogus Cubuk, Barret Zoph, Dandelion Man{\'{e}}, Vijay Vasudevan, and
  Quoc~V. Le.
\newblock Autoaugment: Learning augmentation policies from data.
\newblock In \emph{CVPR}, 2019.

\bibitem[Deng et~al.(2009)Deng, Dong, Socher, Li, Li, and
  Fei-Fei]{deng2009imagenet}
Jia Deng, Wei Dong, Richard Socher, Li-Jia Li, Kai Li, and Li~Fei-Fei.
\newblock {ImageNet}: A large-scale hierarchical image database.
\newblock In \emph{CVPR}, 2009.

\bibitem[Devlin et~al.(2019)Devlin, Chang, Lee, and
  Toutanova]{DBLP:conf/naacl/DevlinCLT19}
Jacob Devlin, Ming{-}Wei Chang, Kenton Lee, and Kristina Toutanova.
\newblock {BERT:} pre-training of deep bidirectional transformers for language
  understanding.
\newblock In \emph{NAACL}, 2019.

\bibitem[Everingham et~al.(2015)Everingham, Eslami, Gool, Williams, Winn, and
  Zisserman]{DBLP:journals/ijcv/EveringhamEGWWZ15}
Mark Everingham, S.~M.~Ali Eslami, Luc~Van Gool, Christopher K.~I. Williams,
  John~M. Winn, and Andrew Zisserman.
\newblock The {PASCAL} visual object classes challenge: {A} retrospective.
\newblock \emph{IJCV}, 2015.

\bibitem[Frankle \& Carbin(2019)Frankle and Carbin]{DBLP:conf/iclr/FrankleC19}
Jonathan Frankle and Michael Carbin.
\newblock The lottery ticket hypothesis: Finding sparse, trainable neural
  networks.
\newblock In \emph{ICLR}, 2019.

\bibitem[Girshick(2015)]{DBLP:conf/iccv/Girshick15}
Ross~B. Girshick.
\newblock Fast {R-CNN}.
\newblock In \emph{ICCV}, 2015.

\bibitem[Gordon et~al.(2018)Gordon, Eban, Nachum, Chen, Wu, Yang, and
  Choi]{DBLP:conf/cvpr/GordonENCWYC18}
Ariel Gordon, Elad Eban, Ofir Nachum, Bo~Chen, Hao Wu, Tien{-}Ju Yang, and
  Edward Choi.
\newblock {MorphNet}: Fast {\&} simple resource-constrained structure learning
  of deep networks.
\newblock In \emph{CVPR}, 2018.

\bibitem[Goyal et~al.(2017)Goyal, Doll{\'{a}}r, Girshick, Noordhuis,
  Wesolowski, Kyrola, Tulloch, Jia, and He]{DBLP:journals/corr/GoyalDGNWKTJH17}
Priya Goyal, Piotr Doll{\'{a}}r, Ross~B. Girshick, Pieter Noordhuis, Lukasz
  Wesolowski, Aapo Kyrola, Andrew Tulloch, Yangqing Jia, and Kaiming He.
\newblock Accurate, large minibatch {SGD:} training imagenet in 1 hour.
\newblock \emph{arXiv:1706.02677}, 2017.

\bibitem[Gross \& Wilber(2016)Gross and Wilber]{gross2016training}
Sam Gross and Michael Wilber.
\newblock Training and investigating residual nets.
\newblock \emph{\url{http://torch.ch/blog/2016/02/04/resnets.html}}, 2016.

\bibitem[Guo et~al.(2016)Guo, Yao, and Chen]{DBLP:conf/nips/GuoYC16}
Yiwen Guo, Anbang Yao, and Yurong Chen.
\newblock Dynamic network surgery for efficient {DNNs}.
\newblock In \emph{NeurIPS}, 2016.

\bibitem[Han et~al.(2015)Han, Pool, Tran, and
  Dally]{DBLP:journals/corr/HanPTD15}
Song Han, Jeff Pool, John Tran, and William~J. Dally.
\newblock Learning both weights and connections for efficient neural networks.
\newblock In \emph{NeurIPS}, 2015.

\bibitem[Han et~al.(2016)Han, Mao, and Dally]{DBLP:journals/corr/HanMD15}
Song Han, Huizi Mao, and William~J. Dally.
\newblock Deep compression: Compressing deep neural network with pruning,
  trained quantization and huffman coding.
\newblock In \emph{ICLR}, 2016.

\bibitem[Hariharan et~al.(2011)Hariharan, Arbelaez, Bourdev, Maji, and
  Malik]{DBLP:conf/iccv/HariharanABMM11}
Bharath Hariharan, Pablo Arbelaez, Lubomir~D. Bourdev, Subhransu Maji, and
  Jitendra Malik.
\newblock Semantic contours from inverse detectors.
\newblock In \emph{ICCV}, 2011.

\bibitem[He et~al.(2016)He, Zhang, Ren, and Sun]{DBLP:conf/cvpr/HeZRS16}
Kaiming He, Xiangyu Zhang, Shaoqing Ren, and Jian Sun.
\newblock Deep residual learning for image recognition.
\newblock In \emph{CVPR}, 2016.

\bibitem[He et~al.(2018)He, Kang, Dong, Fu, and Yang]{DBLP:conf/ijcai/HeKDFY18}
Yang He, Guoliang Kang, Xuanyi Dong, Yanwei Fu, and Yi~Yang.
\newblock Soft filter pruning for accelerating deep convolutional neural
  networks.
\newblock \emph{IJCAI}, 2018.

\bibitem[Hinton et~al.(2015)Hinton, Vinyals, and
  Dean]{DBLP:journals/corr/HintonVD15}
Geoffrey~E. Hinton, Oriol Vinyals, and Jeffrey Dean.
\newblock Distilling the knowledge in a neural network.
\newblock In \emph{NeurIPS Deep Learning and Representation Learning Workshop},
  2015.

\bibitem[Hochreiter \& Schmidhuber(1997)Hochreiter and Schmidhuber]{LSTM}
Sepp Hochreiter and Jurgen Schmidhuber.
\newblock Long short-term memory.
\newblock \emph{Neural Computation}, 1997.

\bibitem[Howard et~al.(2017)Howard, Zhu, Chen, Kalenichenko, Wang, Weyand,
  Andreetto, and Adam]{DBLP:journals/corr/HowardZCKWWAA17}
Andrew~G. Howard, Menglong Zhu, Bo~Chen, Dmitry Kalenichenko, Weijun Wang,
  Tobias Weyand, Marco Andreetto, and Hartwig Adam.
\newblock Mobilenets: Efficient convolutional neural networks for mobile vision
  applications.
\newblock \emph{arXiv:1704.04861}, 2017.

\bibitem[Hu et~al.(2018)Hu, Shen, and Sun]{DBLP:conf/cvpr/HuSS18}
Jie Hu, Li~Shen, and Gang Sun.
\newblock Squeeze-and-excitation networks.
\newblock In \emph{CVPR}, 2018.

\bibitem[Huang et~al.(2016)Huang, Sun, Liu, Sedra, and
  Weinberger]{DBLP:conf/eccv/HuangSLSW16}
Gao Huang, Yu~Sun, Zhuang Liu, Daniel Sedra, and Kilian~Q. Weinberger.
\newblock Deep networks with stochastic depth.
\newblock In \emph{ECCV}, 2016.

\bibitem[Huang et~al.(2017)Huang, Liu, van~der Maaten, and
  Weinberger]{DBLP:conf/cvpr/HuangLMW17}
Gao Huang, Zhuang Liu, Laurens van~der Maaten, and Kilian~Q. Weinberger.
\newblock Densely connected convolutional networks.
\newblock In \emph{CVPR}, 2017.

\bibitem[Huang et~al.(2018)Huang, Liu, van~der Maaten, and
  Weinberger]{DBLP:journals/corr/abs-1711-09224}
Gao Huang, Shichen Liu, Laurens van~der Maaten, and Kilian~Q. Weinberger.
\newblock {CondenseNet}: An efficient {DenseNet} using learned group
  convolutions.
\newblock In \emph{CVPR}, 2018.

\bibitem[Hubara et~al.(2016)Hubara, Courbariaux, Soudry, El{-}Yaniv, and
  Bengio]{DBLP:conf/nips/HubaraCSEB16}
Itay Hubara, Matthieu Courbariaux, Daniel Soudry, Ran El{-}Yaniv, and Yoshua
  Bengio.
\newblock Binarized neural networks.
\newblock In \emph{NeurIPS}, 2016.

\bibitem[Iandola et~al.(2016)Iandola, Moskewicz, Ashraf, Han, Dally, and
  Keutzer]{DBLP:journals/corr/IandolaMAHDK16}
Forrest~N. Iandola, Matthew~W. Moskewicz, Khalid Ashraf, Song Han, William~J.
  Dally, and Kurt Keutzer.
\newblock {SqueezeNet}: {AlexNet}-level accuracy with 50x fewer parameters and
  {\textless}{1MB} model size.
\newblock \emph{arXiv:1602.07360}, 2016.

\bibitem[Ioffe \& Szegedy(2015)Ioffe and Szegedy]{DBLP:conf/icml/IoffeS15}
Sergey Ioffe and Christian Szegedy.
\newblock Batch normalization: Accelerating deep network training by reducing
  internal covariate shift.
\newblock In \emph{ICML}, 2015.

\bibitem[Jang et~al.(2017)Jang, Gu, and Poole]{DBLP:conf/iclr/JangGP17}
Eric Jang, Shixiang Gu, and Ben Poole.
\newblock Categorical reparameterization with gumbel-softmax.
\newblock In \emph{ICLR}, 2017.

\bibitem[Krizhevsky et~al.(2012)Krizhevsky, Sutskever, and
  Hinton]{krizhevsky2012imagenet}
Alex Krizhevsky, Ilya Sutskever, and Geoffrey~E Hinton.
\newblock {ImageNet} classification with deep convolutional neural networks.
\newblock In \emph{NeurIPS}, 2012.

\bibitem[Krizhevsky et~al.(2014)Krizhevsky, Nair, and
  Hinton]{krizhevsky2014cifar}
Alex Krizhevsky, Vinod Nair, and Geoffrey Hinton.
\newblock The {CIFAR}-10 dataset.
\newblock \emph{\url{http://www.cs.toronto.edu/~kriz/cifar.html}}, 2014.

\bibitem[Lemaire et~al.(2019)Lemaire, Achkar, and
  Jodoin]{DBLP:conf/cvpr/LemaireAJ19}
Carl Lemaire, Andrew Achkar, and Pierre{-}Marc Jodoin.
\newblock Structured pruning of neural networks with budget-aware
  regularization.
\newblock In \emph{CVPR}, 2019.

\bibitem[Li et~al.(2017)Li, Kadav, Durdanovic, Samet, and
  Graf]{DBLP:conf/iclr/0022KDSG17}
Hao Li, Asim Kadav, Igor Durdanovic, Hanan Samet, and Hans~Peter Graf.
\newblock Pruning filters for efficient {ConvNets}.
\newblock In \emph{ICLR}, 2017.

\bibitem[Li \& Talwalkar(2019)Li and Talwalkar]{DBLP:conf/uai/LiT19}
Liam Li and Ameet Talwalkar.
\newblock Random search and reproducibility for neural architecture search.
\newblock In \emph{UAI}, 2019.

\bibitem[Liebenwein et~al.(2020)Liebenwein, Baykal, Lang, Feldman, and
  Rus]{DBLP:conf/iclr/LiebenweinBLFR20}
Lucas Liebenwein, Cenk Baykal, Harry Lang, Dan Feldman, and Daniela Rus.
\newblock Provable filter pruning for efficient neural networks.
\newblock In \emph{ICLR}, 2020.

\bibitem[Lin et~al.(2013)Lin, Chen, and Yan]{DBLP:journals/corr/LinCY13}
Min Lin, Qiang Chen, and Shuicheng Yan.
\newblock Network in network.
\newblock \emph{arXiv:1312.4400}, 2013.

\bibitem[Liu et~al.(2019)Liu, Simonyan, and Yang]{liu2018darts}
Hanxiao Liu, Karen Simonyan, and Yiming Yang.
\newblock Darts: Differentiable architecture search.
\newblock In \emph{ICLR}, 2019.

\bibitem[Liu et~al.(2016)Liu, Anguelov, Erhan, Szegedy, Reed, Fu, and
  Berg]{DBLP:conf/eccv/LiuAESRFB16}
Wei Liu, Dragomir Anguelov, Dumitru Erhan, Christian Szegedy, Scott~E. Reed,
  Cheng{-}Yang Fu, and Alexander~C. Berg.
\newblock {SSD:} single shot multibox detector.
\newblock In \emph{ECCV}, 2016.

\bibitem[Long et~al.(2015)Long, Shelhamer, and
  Darrell]{DBLP:conf/cvpr/LongSD15}
Jonathan Long, Evan Shelhamer, and Trevor Darrell.
\newblock Fully convolutional networks for semantic segmentation.
\newblock In \emph{CVPR}, 2015.

\bibitem[Louizos et~al.(2018)Louizos, Welling, and
  Kingma]{DBLP:journals/corr/abs-1712-01312}
Christos Louizos, Max Welling, and Diederik~P. Kingma.
\newblock Learning sparse neural networks through l\({}_{\mbox{0}}\)
  regularization.
\newblock In \emph{ICLR}, 2018.

\bibitem[Luo et~al.(2017)Luo, Wu, and Lin]{DBLP:conf/iccv/LuoWL17}
Jian{-}Hao Luo, Jianxin Wu, and Weiyao Lin.
\newblock {ThiNet}: {A} filter level pruning method for deep neural network
  compression.
\newblock In \emph{ICCV}, 2017.

\bibitem[Luo et~al.(2018)Luo, Tian, Qin, Chen, and Liu]{luo2018neural}
Renqian Luo, Fei Tian, Tao Qin, Enhong Chen, and Tie-Yan Liu.
\newblock Neural architecture optimization.
\newblock In \emph{NeurIPS}, 2018.

\bibitem[Marcus et~al.(1993)Marcus, Santorini, and
  Marcinkiewicz]{DBLP:journals/coling/MarcusSM94}
Mitchell~P. Marcus, Beatrice Santorini, and Mary~Ann Marcinkiewicz.
\newblock Building a large annotated corpus of english: The penn treebank.
\newblock \emph{Computational Linguistics}, 1993.

\bibitem[Molchanov et~al.(2017)Molchanov, Ashukha, and
  Vetrov]{DBLP:conf/icml/MolchanovAV17}
Dmitry Molchanov, Arsenii Ashukha, and Dmitry~P. Vetrov.
\newblock Variational dropout sparsifies deep neural networks.
\newblock In \emph{ICML}, 2017.

\bibitem[Pham et~al.(2018)Pham, Guan, Zoph, Le, and
  Dean]{DBLP:conf/icml/PhamGZLD18}
Hieu Pham, Melody~Y. Guan, Barret Zoph, Quoc~V. Le, and Jeff Dean.
\newblock Efficient neural architecture search via parameter sharing.
\newblock In \emph{ICML}, 2018.

\bibitem[Radosavovic et~al.(2019)Radosavovic, Johnson, Xie, Lo, and
  Doll{\'{a}}r]{DBLP:conf/iccv/Radosavovic0XLD19}
Ilija Radosavovic, Justin Johnson, Saining Xie, Wan{-}Yen Lo, and Piotr
  Doll{\'{a}}r.
\newblock On network design spaces for visual recognition.
\newblock In \emph{ICCV}, 2019.

\bibitem[Rastegari et~al.(2016)Rastegari, Ordonez, Redmon, and
  Farhadi]{DBLP:conf/eccv/RastegariORF16}
Mohammad Rastegari, Vicente Ordonez, Joseph Redmon, and Ali Farhadi.
\newblock {XNOR-Net}: Imagenet classification using binary convolutional neural
  networks.
\newblock In \emph{ECCV}, 2016.

\bibitem[Real et~al.(2019)Real, Aggarwal, Huang, and
  Le]{DBLP:conf/aaai/RealAHL19}
Esteban Real, Alok Aggarwal, Yanping Huang, and Quoc~V. Le.
\newblock Regularized evolution for image classifier architecture search.
\newblock In \emph{AAAI}, 2019.

\bibitem[Savarese \& Maire(2019)Savarese and Maire]{implicitrecurrent}
Pedro Savarese and Michael Maire.
\newblock Learning implicitly recurrent {CNN}s through parameter sharing.
\newblock In \emph{ICLR}, 2019.

\bibitem[Savarese et~al.(2020)Savarese, Silva, and
  Maire]{DBLP:conf/nips/SavareseSM20}
Pedro Savarese, Hugo Silva, and Michael Maire.
\newblock Winning the lottery with continuous sparsification.
\newblock In \emph{NeurIPS}, 2020.

\bibitem[Sifre \& Mallat(2014)Sifre and Mallat]{sifre2014rigid}
Laurent Sifre and PS~Mallat.
\newblock \emph{Rigid-motion scattering for image classification}.
\newblock PhD thesis, Ecole Polytechnique, CMAP, 2014.

\bibitem[Simonyan \& Zisserman(2015)Simonyan and
  Zisserman]{DBLP:journals/corr/SimonyanZ14a}
Karen Simonyan and Andrew Zisserman.
\newblock Very deep convolutional networks for large-scale image recognition.
\newblock In \emph{ICLR}, 2015.

\bibitem[Tan \& Le(2019)Tan and Le]{DBLP:conf/icml/TanL19}
Mingxing Tan and Quoc~V. Le.
\newblock Efficientnet: Rethinking model scaling for convolutional neural
  networks.
\newblock In \emph{ICML}, 2019.

\bibitem[Tan et~al.(2019)Tan, Chen, Pang, Vasudevan, Sandler, Howard, and
  Le]{tan2019mnasnet}
Mingxing Tan, Bo~Chen, Ruoming Pang, Vijay Vasudevan, Mark Sandler, Andrew
  Howard, and Quoc~V Le.
\newblock Mnasnet: Platform-aware neural architecture search for mobile.
\newblock In \emph{CVPR}, 2019.

\bibitem[Vaswani et~al.(2017)Vaswani, Shazeer, Parmar, Uszkoreit, Jones, Gomez,
  Kaiser, and Polosukhin]{vaswani2017attention}
Ashish Vaswani, Noam Shazeer, Niki Parmar, Jakob Uszkoreit, Llion Jones,
  Aidan~N Gomez, {\L}ukasz Kaiser, and Illia Polosukhin.
\newblock Attention is all you need.
\newblock In \emph{NeurIPS}, 2017.

\bibitem[Wei et~al.(2016)Wei, Wang, Rui, and Chen]{DBLP:conf/icml/WeiWRC16}
Tao Wei, Changhu Wang, Yong Rui, and Chang~Wen Chen.
\newblock Network morphism.
\newblock In \emph{ICML}, 2016.

\bibitem[Wen et~al.(2018)Wen, He, Rajbhandari, Zhang, Wang, Liu, Hu, Chen, and
  Li]{DBLP:conf/iclr/WenHRZWLHC018}
Wei Wen, Yuxiong He, Samyam Rajbhandari, Minjia Zhang, Wenhan Wang, Fang Liu,
  Bin Hu, Yiran Chen, and Hai Li.
\newblock Learning intrinsic sparse structures within long short-term memory.
\newblock In \emph{ICLR}, 2018.

\bibitem[Wen et~al.(2020)Wen, Yan, Chen, and Li]{DBLP:conf/kdd/Wen0CL20}
Wei Wen, Feng Yan, Yiran Chen, and Hai Li.
\newblock Autogrow: Automatic layer growing in deep convolutional networks.
\newblock In \emph{KDD}, 2020.

\bibitem[Wortsman et~al.(2019)Wortsman, Farhadi, and
  Rastegari]{DBLP:conf/nips/WortsmanFR19}
Mitchell Wortsman, Ali Farhadi, and Mohammad Rastegari.
\newblock Discovering neural wirings.
\newblock In \emph{NeurIPS}, 2019.

\bibitem[Wu et~al.(2019)Wu, Dai, Zhang, Wang, Sun, Wu, Tian, Vajda, Jia, and
  Keutzer]{DBLP:conf/cvpr/WuDZWSWTVJK19}
Bichen Wu, Xiaoliang Dai, Peizhao Zhang, Yanghan Wang, Fei Sun, Yiming Wu,
  Yuandong Tian, Peter Vajda, Yangqing Jia, and Kurt Keutzer.
\newblock Fbnet: Hardware-aware efficient convnet design via differentiable
  neural architecture search.
\newblock In \emph{CVPR}, 2019.

\bibitem[Xie \& Yuille(2017)Xie and Yuille]{xie2017genetic}
Lingxi Xie and Alan~L Yuille.
\newblock Genetic cnn.
\newblock In \emph{ICCV}, 2017.

\bibitem[Xie et~al.(2019{\natexlab{a}})Xie, Kirillov, Girshick, and
  He]{DBLP:conf/iccv/XieKGH19}
Saining Xie, Alexander Kirillov, Ross~B. Girshick, and Kaiming He.
\newblock Exploring randomly wired neural networks for image recognition.
\newblock In \emph{ICCV}, 2019{\natexlab{a}}.

\bibitem[Xie et~al.(2019{\natexlab{b}})Xie, Zheng, Liu, and
  Lin]{DBLP:conf/iclr/XieZLL19}
Sirui Xie, Hehui Zheng, Chunxiao Liu, and Liang Lin.
\newblock {SNAS:} stochastic neural architecture search.
\newblock In \emph{ICLR}, 2019{\natexlab{b}}.

\bibitem[Xie et~al.(2020)Xie, Zhang, Zhu, Huang, and
  Lin]{DBLP:conf/eccv/XieZZHL20}
Zhenda Xie, Zheng Zhang, Xizhou Zhu, Gao Huang, and Stephen Lin.
\newblock Spatially adaptive inference with stochastic feature sampling and
  interpolation.
\newblock In \emph{ECCV}, 2020.

\bibitem[Yang et~al.(2018)Yang, Howard, Chen, Zhang, Go, Sandler, Sze, and
  Adam]{DBLP:conf/eccv/YangHCZGSSA18}
Tien{-}Ju Yang, Andrew~G. Howard, Bo~Chen, Xiao Zhang, Alec Go, Mark Sandler,
  Vivienne Sze, and Hartwig Adam.
\newblock {NetAdapt}: Platform-aware neural network adaptation for mobile
  applications.
\newblock In \emph{ECCV}, 2018.

\bibitem[Yu et~al.(2020)Yu, Sciuto, Jaggi, Musat, and
  Salzmann]{DBLP:conf/iclr/YuSJMS20}
Kaicheng Yu, Christian Sciuto, Martin Jaggi, Claudiu Musat, and Mathieu
  Salzmann.
\newblock Evaluating the search phase of neural architecture search.
\newblock In \emph{ICLR}, 2020.

\bibitem[Zagoruyko \& Komodakis(2016)Zagoruyko and
  Komodakis]{DBLP:conf/bmvc/ZagoruykoK16}
Sergey Zagoruyko and Nikos Komodakis.
\newblock Wide residual networks.
\newblock In \emph{BMVC}, 2016.

\bibitem[Zaremba et~al.(2014)Zaremba, Sutskever, and
  Vinyals]{DBLP:journals/corr/ZarembaSV14}
Wojciech Zaremba, Ilya Sutskever, and Oriol Vinyals.
\newblock Recurrent neural network regularization.
\newblock \emph{arXiv:1409.2329}, 2014.

\bibitem[Zhang et~al.(2018)Zhang, Zhou, Lin, and
  Sun]{DBLP:journals/corr/ZhangZLS17}
Xiangyu Zhang, Xinyu Zhou, Mengxiao Lin, and Jian Sun.
\newblock {ShuffleNet}: An extremely efficient convolutional neural network for
  mobile devices.
\newblock In \emph{CVPR}, 2018.

\bibitem[Zoph \& Le(2017)Zoph and Le]{DBLP:conf/iclr/ZophL17}
Barret Zoph and Quoc~V. Le.
\newblock Neural architecture search with reinforcement learning.
\newblock In \emph{ICLR}, 2017.

\bibitem[Zoph et~al.(2018)Zoph, Vasudevan, Shlens, and Le]{zoph2017learning}
Barret Zoph, Vijay Vasudevan, Jonathon Shlens, and Quoc~V Le.
\newblock Learning transferable architectures for scalable image recognition.
\newblock In \emph{CVPR}, 2018.

\end{thebibliography}
\bibliographystyle{iclr2021_conference}

\newpage
\appendix
\section{Appendix}
\label{sec:appendix}

\subsection{More detailed analysis for Budget-aware Growing} \label{append:buget_grow}
Conducting grid search on trade-off parameters $\lambda_1$ and $\lambda_2$ is prohibitively laborious and time-consuming. For example, to grow an efficient network on CIFAR-10, one needs to repeat many times a run of 160-epochs training, and then pick the best model from all grown candidates. 
To avoid this tedious iterative process, instead of using constants $\lambda_1$ and $\lambda_2$, we dynamically update $\lambda_1$ and $\lambda_2$ in our one-shot budget-aware growing optimization. 

Here we discuss about how budget-aware dynamic growing works in our method.
Without loss of generality, we derive the $m_c$'s SGD update rule for the $\ell_0$ regularization term in Eq.~\ref{loss} as: 
\begin{eqnarray}
m_c \leftarrow m_c - \eta \lambda_1^{base}\Delta u \frac{\delta \ell}{\delta m_c} - \eta \mu\lambda_1^{base}\Delta u m_c
\end{eqnarray}
where $\eta$ is the learning rate and $\mu$ is the weight decay factor. At the beginning of growing epochs, when the architecture is very over-sparsified, $\Delta u$ and $\lambda_1^{base}\Delta u$ are negative values. Then $m_c$'s update is along the \textbf{opposite} direction of the $\ell_0$ regularization term's gradients, encouraging $m_c$'s sparsification. As a result, some zero-valued $m_c$ will be activated and the model complexity is strongly increased to acquire enough capacity for successful training. Then, growing becomes gradually weaker as the network's sparsity approaches the budget ($\Delta u$ to zero). Note that if the architecture is over-parameterized, $\Delta u$ and $\lambda_1^{base}\Delta u$ become positive and SGD's update rule is the same as that of $\ell_0$ regularization. As such, our budget-aware growing can automatically and dynamically adapt the architecture complexity not only based on the task loss $L_E$ but also on the practical budget requirements in the one-shot training process.

We also note that NAS methods usually use the validation accuracy as a target during their architecture optimization phase, which may require some prior knowledge of validation accuracy on a given dataset.
Our growing procedure chooses sparsity budget instead of accuracy as target because: (1) During growing, validation accuracy is influenced not only by architectures but also model weights. Directly using $\Delta acc$ may lead to sub-optimal architecture optimization. (2) A sparsity budget target is more practical and easier to set according to target devices for deployment.

\subsection{Details of Evaluation Datasets} \label{append_dataset}
Evaluation is conducted on various tasks to demonstrate the effectiveness of our proposed method.
For image classification, we use CIFAR-10~\citep{krizhevsky2014cifar} and ImageNet~\citep{deng2009imagenet}:
CIFAR-10 consists of 60,000 images of 10 classes, with 6,000 images per class. The train and test sets contain 50,000 and 10,000 images respectively. ImageNet is a large dataset for visual recognition which contains over 1.2M images in the training set and 50K images in the validation set covering 1,000 categories. For semantic segmentation, we use the PASCAL VOC 2012~\citep{DBLP:journals/ijcv/EveringhamEGWWZ15} benchmark which contains 20 foreground object classes and one background class. The original dataset contains 1,464 (train), 1,449 (val), and 1,456 (test) pixel-level labeled images for training, validation, and testing, respectively. The dataset is augmented by the extra annotations provided by~\citep{DBLP:conf/iccv/HariharanABMM11}, resulting in 10,582 training images. For language modeling, we use the word level Penn Treebank (PTB) dataset~\citep{DBLP:journals/coling/MarcusSM94} which consists of 929k training words, 73k validation words, and 82k test words, with 10,000 unique words in its vocabulary. 
\subsection{Unpruned Baseline Models} \label{unprune_model}
For CIFAR-10, we use VGG-16~\citep{DBLP:journals/corr/SimonyanZ14a} with BatchNorm~\citep{DBLP:conf/icml/IoffeS15}, ResNet-20~\citep{DBLP:conf/cvpr/HeZRS16} and WideResNet-28-10~\citep{DBLP:conf/bmvc/ZagoruykoK16} as baselines. We adopt a standard data augmentation scheme (shifting/mirroring) following~\citep{DBLP:journals/corr/LinCY13, DBLP:conf/eccv/HuangSLSW16}, and normalize the input data with channel means and standard deviations.
Note that we use the CIFAR version of ResNet-20\footnote{\text{https://github.com/akamaster/pytorch\_resnet\_cifar10/blob/master/resnet.py}},
VGG-16\footnote{https://github.com/kuangliu/pytorch-cifar/blob/master/models/vgg.py}, 
and WideResNet-28-10\footnote{https://github.com/meliketoy/wide-resnet.pytorch/blob/master/networks/wide\_resnet.py}.
VGG-16, ResNet-20, and WideResNet-28-10 are trained for 160, 160, and 200 epochs, respectively, with a batch size of 128 and initial learning rate of 0.1. For VGG-16 and ResNet-20, we divide learning rate by 10 at epochs 80 and 120, and set the weight decay and momentum as $10^{-4}$ and 0.9. For WideResNet-28-10, the learning rate is divided by 5 at epochs 60, 120, and 160; the weight decay and momentum are set to $5\times10^{-4}$ and 0.9.
For ImageNet, we train the baseline ResNet-50 and MobileNetV1 models following the respective papers. We adopt the same data augmentation scheme as in~\citep{gross2016training} and report top-1 validation accuracy. 
For semantic segmentation, the performance is measured in terms of pixel intersection-over-union (IOU) averaged across the 21 classes (mIOU). We use Deeplab-v3-ResNet-101\footnote{https://github.com/chenxi116/DeepLabv3.pytorch}~\citep{DBLP:journals/corr/ChenPSA17} as the baseline model following the training details in~\citep{DBLP:journals/corr/ChenPSA17}.
For language modeling, we use vanilla two-layer stacked LSTM~\citep{DBLP:journals/corr/ZarembaSV14} as a baseline. The dropout keep ratio is 0.35 for the baseline model. The vocabulary size, embedding size, and hidden size of the stacked LSTMs are set as 10,000, 1,500, and 1,500, respectively, which is consistent with the settings in~\citep{DBLP:journals/corr/ZarembaSV14}.

\subsection{MobileNetV1 Channel Growing on ImageNet}\label{mobile_imagenet}
To further validate the effectiveness of the proposed method on compact networks, we grow the filters of MobileNetV1 on ImageNet and compare the performance of our method to the results reported directly in the respective papers, as shown in Table~\ref{tab:mobile_imagenet:result}. In MobileNetV1 experiments, following the same setting with Netadapt~\citep{DBLP:conf/eccv/YangHCZGSSA18}, we apply our method on both (1) small setting: growing MobileNetV1(128) with 0.5 multiplier while setting the original model's multiplier as 0.25 for comparison and (2) large setting: growing standard MobileNetV1(224) while setting the original model's multiplier as 0.75 for comparison. Note that MobileNetV1 is one of the most compact networks, and thus is more challenging to simplify than other larger networks. Our lower-cost growing method can still generate a sparser MobileNetV1 model compared with competing methods.

\begin{table}[h]
\setlength{\tabcolsep}{8pt}
\footnotesize
\begin{center}
\caption{Overview of the pruning performance of each algorithm on MobileNetV1 ImageNet.}
\label{tab:mobile_imagenet:result}
\vspace{1pt}
\begin{tabular}{ccccccc}
\toprule
&\multicolumn{1}{c}{{Model}}
&\multicolumn{1}{c}{{Method}}
&\multicolumn{1}{c}{Top-1 Val Acc(\%)}
&\multicolumn{1}{c}{FLOPs(\%)}
&\multicolumn{1}{c}{Train-Cost Savings($\times$)} \\
\midrule
& &Original(25\%) & 45.1 (+0.0) &100 & 1.0($\times$)\\
&MobileNet &MorphNet &\underline{46.0 (+0.9)} &110 & 0.9($\times$) \\
&V1(128) &Netadapt &\textbf{46.3 (+1.2)} &\underline{81} &\underline{1.1}($\times$)\\
& &Ours &\underline{46.0 (+0.9)} &\textbf{73} &\textbf{1.7}($\times$)\\
\midrule
&\multirow{2}{*}{MobileNet} &Original(75\%) & 68.8 (+0.0)  &100 & 1.0($\times$)\\
&\multirow{2}{*}{V1(224)} &Netadapt &\underline{69.1 (+0.3)}  &\underline{87} & 1.2($\times$) \\
& &Ours &\textbf{69.3 (+0.5)}  &\textbf{83} &\textbf{1.5}($\times$)\\
\bottomrule
\end{tabular}
\end{center}
\vskip -0.2in
\end{table}

\subsection{Deeplab-v3-ResNet-101 on PASCAL VOC 2012}\label{deeplab}
We also test the effectiveness of our proposed method on a semantic segmentation task by growing a Deeplab-v3-ResNet-101 model's filter numbers from scratch directly on the PASCAL VOC 2012 dataset. We apply our method to both the ResNet-101 backbone and ASPP module. Compared to the baseline, the final generated network reduces the FLOPs by 58.5\% and the parameter count by 49.8\%, while approximately maintaining mIoU (76.5\% to 76.4\%). See Table~\ref{tab:seg:result}.
\begin{table}[h]
\footnotesize
\begin{center}
\caption{Results on the PASCAL VOC dataset.}
\label{tab:seg:result}
\vspace{3pt}
\begin{tabular}{ccccccc}
\toprule
&\multicolumn{1}{c}{{Model}}
&\multicolumn{1}{c}{{Method}}
&\multicolumn{1}{c}{mIOU} 
&\multicolumn{1}{c}{Params(M)}
&\multicolumn{1}{c}{FLOPs(\%)} 
&\multicolumn{1}{c}{Train-Cost Savings($\times$)} \\

\midrule
&Deeplab &Original &76.5 (-0.0) &58.0 (100\%) &100 & 1.0($\times$)\\
&-v3- &L1-Pruning &75.1 (-1.4) &45.7 (78.8\%) &62.5 & 1.3($\times$) \\
&ResNet101 &Ours &\textbf{76.4 (-0.1)} &\textbf{29.1 (50.2\%)} &\textbf{41.5} & \textbf{2.3}($\times$) \\
\bottomrule
\end{tabular}
\end{center}
\end{table}

\subsection{Extension to recurrent cells on PTB dataset} \label{lstm_ext}
We focus on LSTMs~\citep{LSTM} with $l_{h}$ hidden neurons, a common variant\footnote{The proposed configuration space can be readily applied to the compression of GRUs~\citep{DBLP:conf/emnlp/ChoMGBBSB14} and vanilla RNNs.} of RNNs that learns long-term dependencies:
\begin{eqnarray} \label{LSTM}
&f_t = \sigma_g((W_{f} \odot (\mathbf{e} m_c^T) ) x_t + (U_{f} \odot (m_cm_c^T)) h_{t-1} + b_f) \nonumber \\
&i_t = \sigma_g((W_{i} \odot (\mathbf{e} m_c^T)) x_t + (U_{i} \odot (m_cm_c^T)) h_{t-1} + b_i)  \nonumber\\
&o_t = \sigma_g((W_{o} \odot (\mathbf{e} m_c^T)) x_t + (U_{o} \odot (m_cm_c^T)) h_{t-1} + b_o) \nonumber \\
&\tilde{c}_t = \sigma_h((W_{c} \odot (\mathbf{e} m_c^T)) x_t + (U_{c} \odot (m_cm_c^T)) h_{t-1} + b_c)  \nonumber \\
&c_t = f_t \odot c_{t-1} + i_t \odot \tilde{c}_t, \quad h_t = o_t \odot \sigma_h(c_t) \quad s.t. \quad m_c \in \{0,1\}^{l_{h}}, \mathbf{e} = {1}^{l_{h}}\,,
\end{eqnarray}
where $\sigma_g$ is the sigmoid function, $\odot$ denotes element-wise multiplication and $\sigma_h$  is  the  hyperbolic  tangent  function. $x_t$ denotes  the  input  vector at  the  time-step $t$, $h_t$ denotes  the  current  hidden  state,  and $c_t$ denotes  the long-term  memory  cell  state.
$W_f, W_i, W_o, W_c$ denote the input-to-hidden weight matrices and $U_f, U_i, U_o, U_c$ denote the hidden-to-hidden weight matrices. $m_c$ is binary indicator and shared across all the gates to control the sparsity of hidden neurons. 

We compare our proposed method with ISS based on vanilla two-layer stacked LSTM. As shown in Table~\ref{tab:ptb:result}, our method finds more compact model structure at lower training cost, while achieving similar perplexity on both validation and test sets. 
These improvements may be due to the fact that our method dynamically grows and prunes the hidden neurons from very simple status towards a better trade-off between model complexity and performance than that of ISS, which simply uses the group lasso to penalize the norms of all groups collectively for compactness.
\begin{table}[h]
\footnotesize
\setlength{\tabcolsep}{4pt}
\begin{center}
\caption{Results on the PTB dataset.}
\label{tab:ptb:result}
\vspace{3pt}
\begin{tabular}{ccccccc}
\toprule
&\multicolumn{1}{c}{{Method}}
&\multicolumn{1}{c}{Perplexity  (val,test)} 
&\multicolumn{1}{c}{Final Structure}
&\multicolumn{1}{c}{Weight(M)} 
&\multicolumn{1}{c}{FLOPs(\%)}
&\multicolumn{1}{c}{Train-Cost Savings($\times$)} \\
\midrule
&Original &(82.57, 78.57) &(1500, 1500) &66.0M (100\%) &100 &1.0($\times$) \\
&ISS &(82.59, \textbf{78.65}) &(373, 315) &21.8M (33.1\%) &13.4 &3.8($\times$) \\
&Ours &(\textbf{82.46}, 78.68) &\textbf{(310, 275)} &\textbf{20.6M (31.2\%)} &\textbf{11.9} &\textbf{5.1}($\times$) \\
\bottomrule
\end{tabular}
\end{center}
\end{table}
\subsection{Variants of Initial Seed Architecture} \label{seed_arch}
In Table~\ref{tab:autogrow:seed}, we make a detailed comparison among initial seed architecture variants of ours and AutoGrow~\citep{DBLP:conf/kdd/Wen0CL20}. For both ours and AutoGrow, ``Basic'' and ``Bottleneck'' refer to ResNets with standard basic and bottleneck residual blocks, while ``PlainLayers'' refers to stacked convolutional, batch normalization, and ReLU layer combinations. Similar with standard ResNets, for variants of the seed architecture, we adopt three stages for CIFAR-10 and four stages for ImageNet. PlainNets can be obtained by simply removing shortcuts from these ResNet seed variants with equal stage numbers. For each stage, we start from only one growing unit, within which initial filter numbers are also initialized at one for channel growing. 

\begin{table}[h]
\footnotesize
\begin{center}
\caption{A detailed comparison among seed architecture variants of our method and AutoGrow~\citep{DBLP:conf/kdd/Wen0CL20}. In growing units term, ``Basic'' and ``Bottleneck'' refer to ResNets with standard basic and bottleneck residual blocks while ``PlainLayers'' refers to standard convolutional layer, BN, and ReLu layer combinations in VGG-like networks without shortcuts.}
\label{tab:autogrow:seed}
\begin{tabular}{@{}ccccccc@{}}
\toprule
\multicolumn{1}{c}{{Families}}
&\multicolumn{1}{c}{Variants}
&\multicolumn{1}{c}{Methods}
&\multicolumn{1}{c}{Channel Growing}
&\multicolumn{1}{c}{Growing Units}
&\multicolumn{1}{c}{Stages}
&\multicolumn{1}{c}{Shortcuts} \\
\cline{1-7}
   &\multirow{2}{*}{\textit{Basic3ResNet}} &Ours &\cmark &\multirow{2}{*}{Basic} &\multirow{2}{*}{3} &\cmark \\
\multirow{2}{*}{ResNet} & &AutoGrow &\xmark & & &\cmark \\
\cline{2-7}
  &\multirow{2}{*}{\textit{Bottleneck4ResNet}} &Ours &\cmark &\multirow{2}{*}{Bottleneck} &\multirow{2}{*}{4} &\cmark \\
 & &AutoGrow &\xmark & & &\cmark \\
\cline{1-7}
&\multirow{2}{*}{\textit{Plain3Net}} &Ours &\cmark &\multirow{2}{*}{PlainLayers} &\multirow{2}{*}{3} &\xmark \\
\multirow{2}{*}{VGG-like} & &AutoGrow &\xmark & & &\xmark\\
\cline{2-7}
  &\multirow{2}{*}{\textit{Plain4Net}} &Ours &\cmark &\multirow{2}{*}{PlainLayers} &\multirow{2}{*}{4} &\xmark \\
 & &AutoGrow &\xmark & & &\xmark \\
\bottomrule
\end{tabular}
\end{center}
\vskip -0.1in
\end{table}

\subsection{Track of Any-time Sparsification during channel growing} \label{sec:append_anytime_sparse}
\begin{figure}[htb]
   \begin{minipage}[t]{0.49\linewidth}
      \centering
      \vspace{0pt}
      \includegraphics[width=\linewidth]{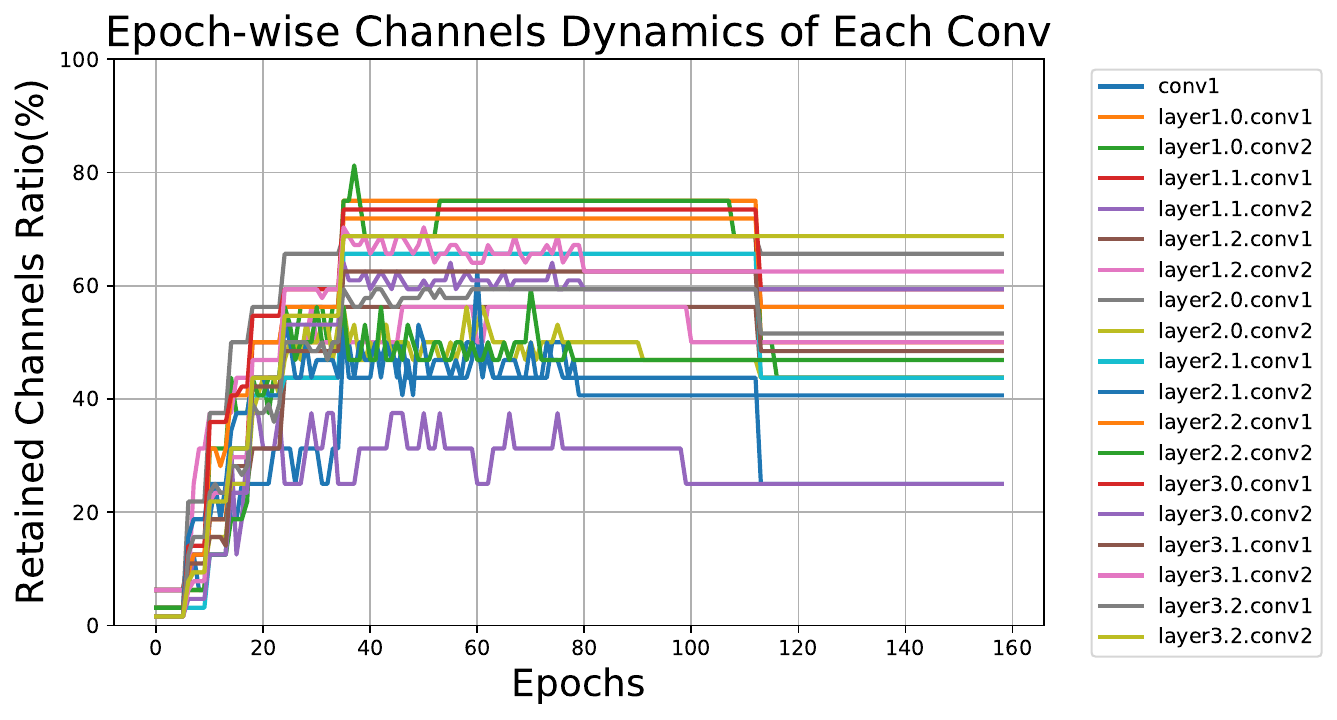}\\
      \vspace{-5pt}
      \caption{Epoch-wise retained channel ratio dynamics for each layer in ResNet-20.}
      \label{fig:avg_channel_res20}
   \end{minipage}
   \hfill
   \begin{minipage}[t]{0.49\linewidth}
      \centering
      \vspace{0pt}
      \includegraphics[width=\linewidth]{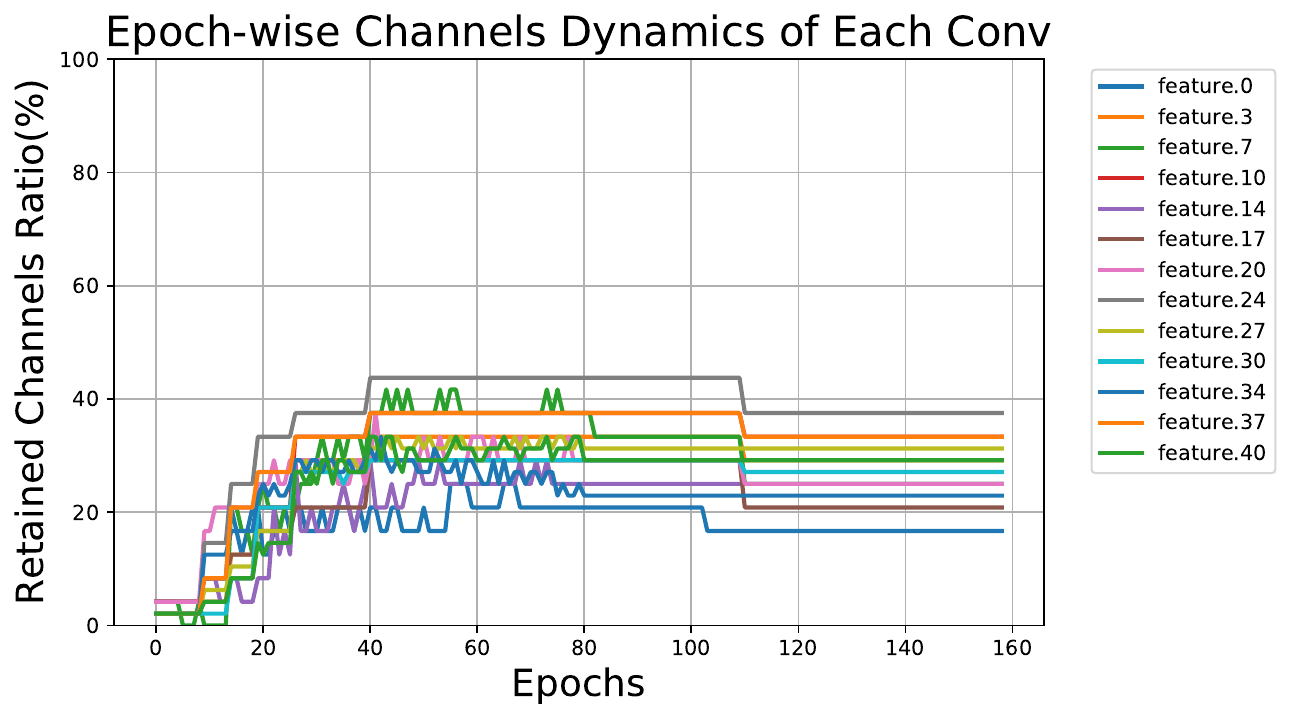}\\
      \vspace{-8pt}
      \caption{Epoch-wise retained channel ratio dynamics for each layer in VGG-16.}
      \label{fig:avg_channel_vgg16}
   \end{minipage}
\end{figure}

\begin{figure}[htb]
   \begin{minipage}[t]{\linewidth}
      \centering
      \vspace{0pt}
      \includegraphics[width=\linewidth]{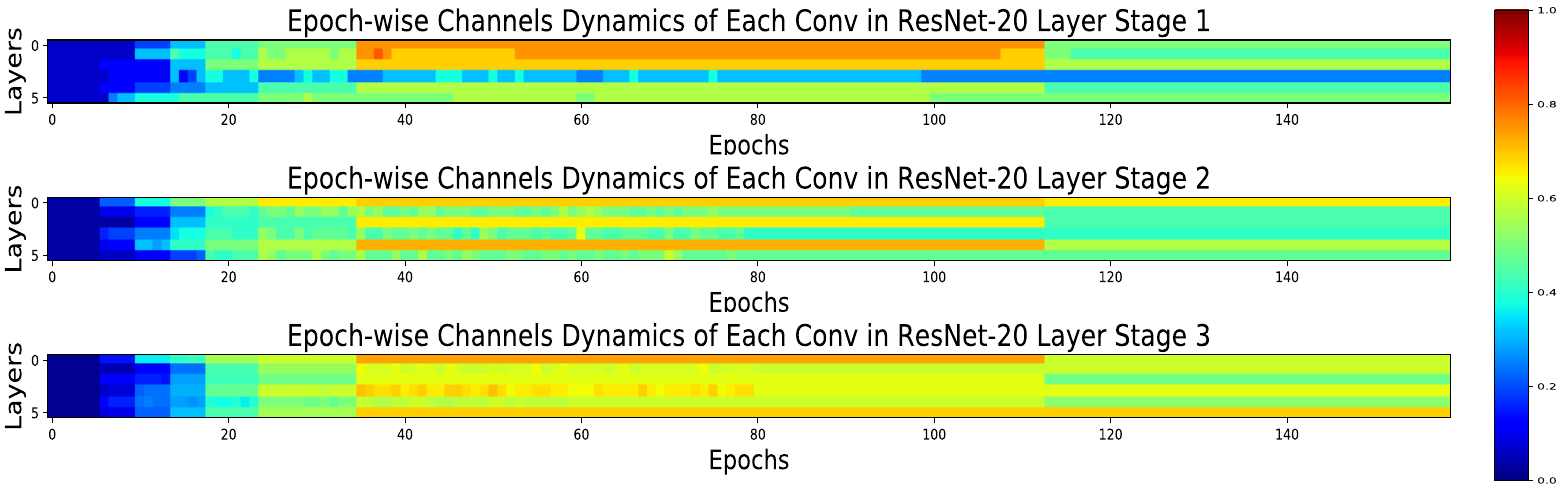}\\
      \vspace{-8pt}
      \caption{Visualization of retained channel ratio dynamics for each stage in ResNet-20.}
      \label{fig:avg_channel_res20_stage}
   \end{minipage}
   ~\\
   \begin{minipage}[t]{\linewidth}
      \centering
      \vspace{0pt}
      \includegraphics[width=\linewidth]{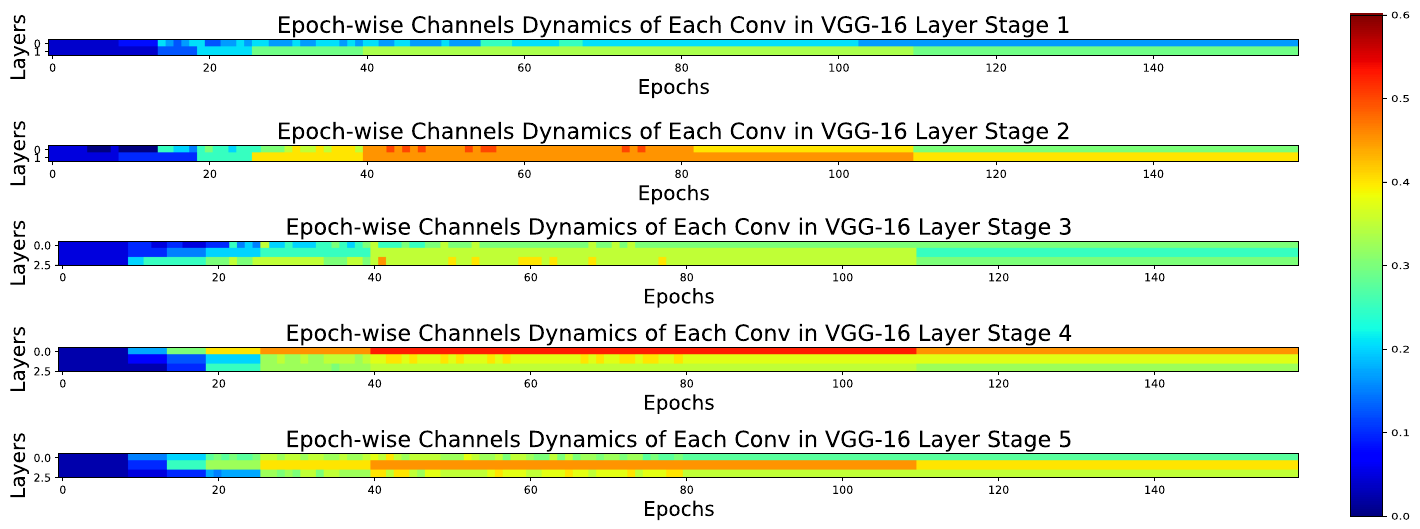}\\
      \vspace{-8pt}
      \caption{Visualization of retained channel ratio dynamics for each stage in VGG-16.}
      \label{fig:avg_channel_vgg16_stage}
   \end{minipage}
\end{figure}

Figure~\ref{fig:avg_channel_res20} and Figure~\ref{fig:avg_channel_vgg16} show the dynamics of train-time growing channel ratios of ResNet-20 and VGG-16 on CIFAR-10, respectively. To better analyze the growing patterns, we visualize the channel dynamics grouped by stages in Figure~\ref{fig:avg_channel_res20_stage} for ResNet-20 and Figure~\ref{fig:avg_channel_vgg16_stage} for VGG-16, respectively. Note that, for VGG-16, we divide it into 5 stages based on the pooling layer positions and normalize channel ratios by 0.5 for better visualization.
We see that our method grows more channels of earlier layers within each stage of ResNet-20.  
Also, the final channel sparsity of ResNet-20 is more uniform due to the residual connections.

\subsection{FLOPs-based Budget-aware Growing} \label{flops_budget_append}

\begin{figure}[tb]
   \begin{minipage}[t]{0.42\linewidth}
      \centering
      \vspace{-1pt}
      \includegraphics[width=\linewidth]{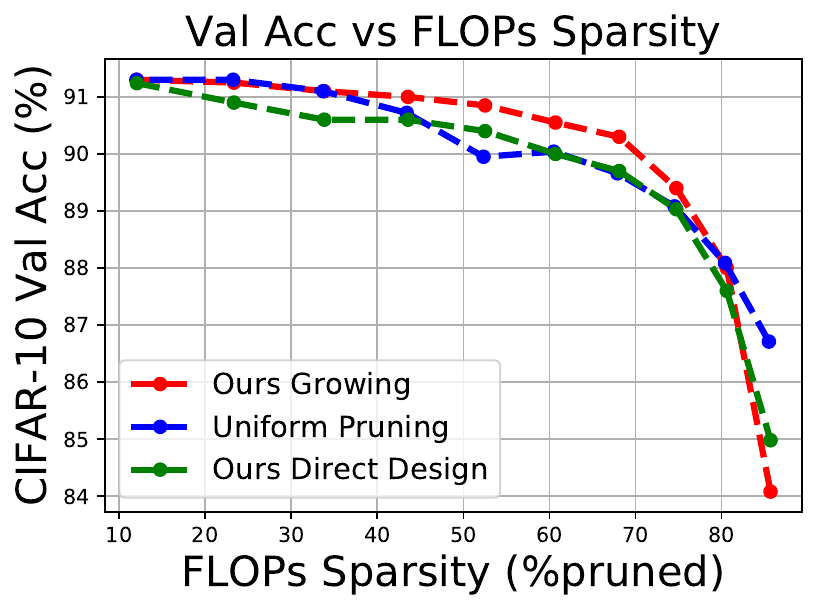}\\
      \vspace{-8pt}
      \caption{Pruned architectures obtained by ablated methods with different FLOPs sparsity.}
      \label{fig:acc_flops}
   \end{minipage}
   \hfill
   \begin{minipage}[t]{0.56\linewidth}
      \centering
      \vspace{0pt}
      \includegraphics[width=\linewidth]{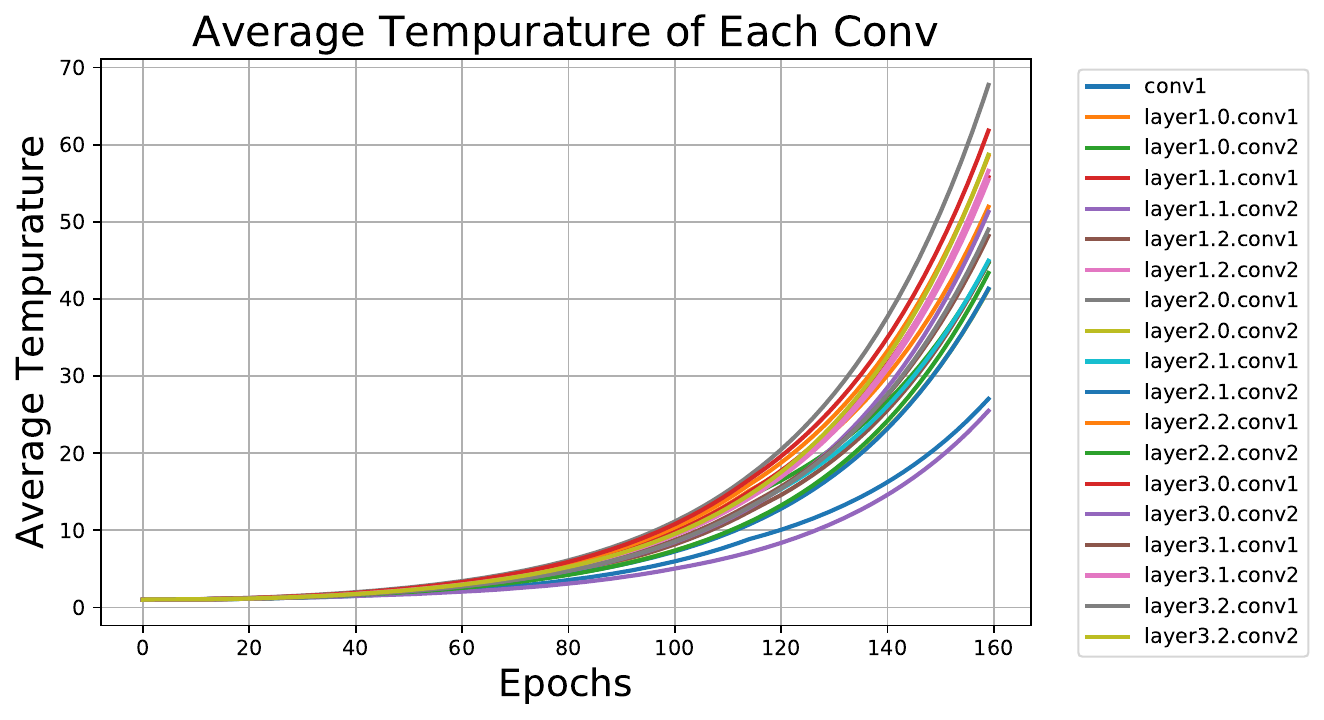}\\
      \vspace{-5pt}
      \caption{Structure-wise separate temperature dynamics in channel growing.}
      \label{fig:temp_vis}
   \end{minipage}
\end{figure}

We also investigate the effectiveness of setting a FLOPs target for budget-aware growing in Figure~\ref{fig:acc_flops}. We observe similar trends among \textit{uniform pruning}, \textit{ours growing}, and \textit{ours direct design}: in most FLOPs budget settings, our growing method outperforms direct design and uniform pruning. We also observe that when setting extreme sparse FLOPs target (\emph{e.g.,}~$85\%$), our method achieves lower accuracy than the other two variants. The reason is that our channel growing is forced to only grow architectures from $\sim 99\%$ sparsity up to $\sim 85\%$ FLOPs and $\sim 90\%$ parameters sparsity, during which models cannot acquire enough capacity to be well trained.

\subsection{Interactions between learning rate and temperature schedulers} \label{interaction_lr_temp}
\begin{figure}[tb]
   \begin{minipage}[t]{0.48\linewidth}
      \centering
      \vspace{0pt}
      \includegraphics[width=\linewidth]{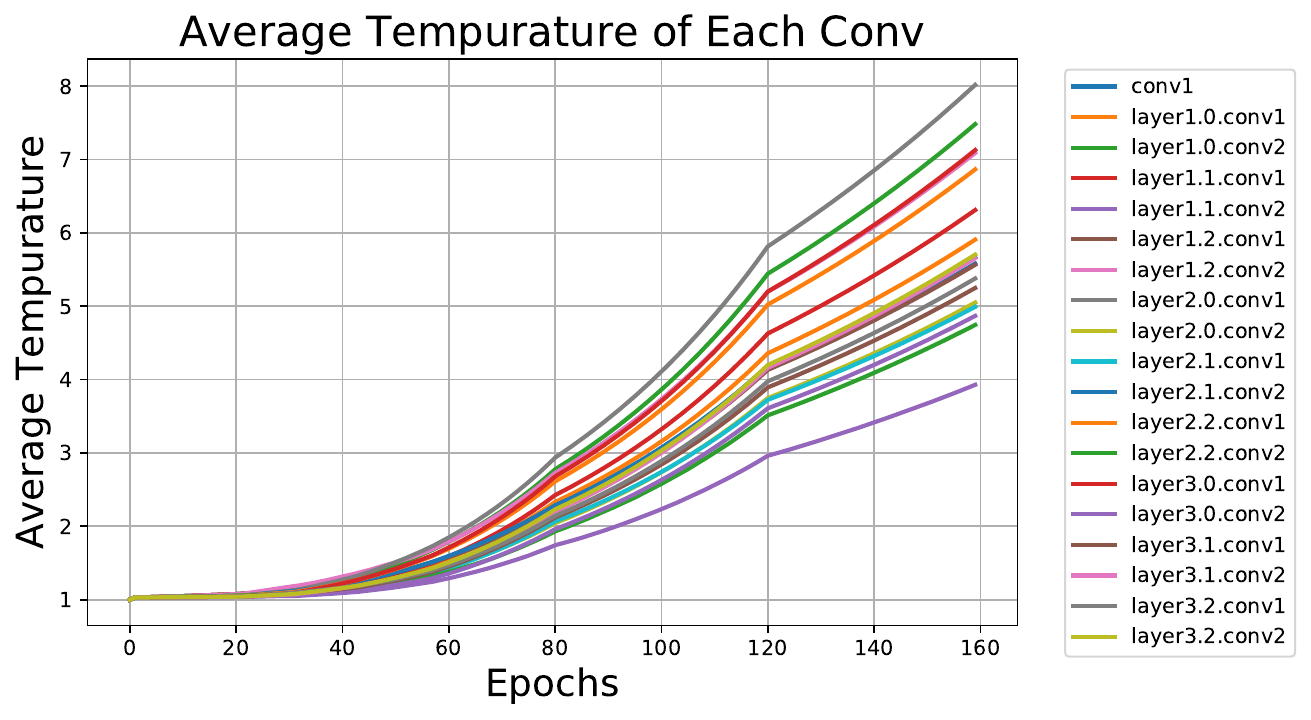}\\
      \vspace{-5pt}
      \caption{Structure-wise separate decayed temperature dynamics in channel growing.}
      \label{fig:temp_vis_decay2}
   \end{minipage}
   \hfill
   \begin{minipage}[t]{0.50\linewidth}
      \centering
      \vspace{0pt}
      \includegraphics[width=\linewidth]{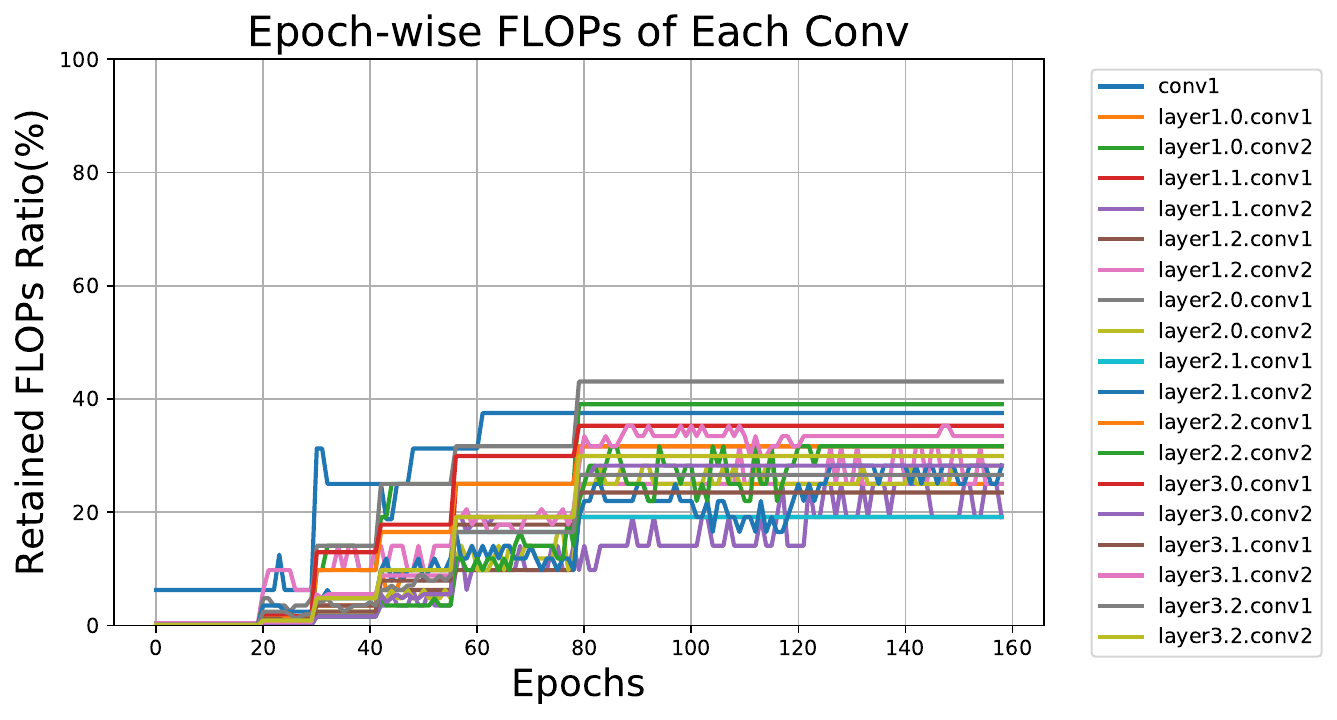}\\
      \vspace{-6pt}
      \caption{Track of epoch-wise train-time FLOPs for channel growing in ResNet-20.}
      \label{fig:res20_track_decay2}
   \end{minipage}
\end{figure}

Two factors influence the growing optimization speed in our method: temperature and learning rate, which are hyperparameters controlled by their respective schedulers. We first visualize the structure-wise separate temperature dynamics in Figure~\ref{fig:temp_vis} by averaging temperatures per layer during ResNet-20 channel growing on CIFAR-10. We see that temperatures are growing with different rates for channels.
Usually, low learning rate and high temperature in late training epochs make the network growing optimization become very stable. In Figure~\ref{fig:temp_vis_decay2}, we deliberately decay $\gamma$ in the temperature scheduler, mirroring the learning rate decay schedule, in order to force growing until the end. As shown in Figure~\ref{fig:res20_track_decay2}, our method is still adapting some layers even at the last epoch. We find that such instability degrades performance, since some newly grown filters may not have enough time to become well trained.

\end{document}